\documentclass[10pt,twoside]{IEEEtran}
\ifCLASSOPTIONcompsoc
  \usepackage[nocompress]{cite}
\else
  \usepackage{cite}
\fi
\ifCLASSINFOpdf
\else
\fi
\usepackage{color}
\usepackage{hyperref}
\usepackage{float}
\usepackage{graphicx,graphics}
\usepackage[table,xcdraw]{xcolor}
\usepackage{amsmath,bm}
\usepackage{amssymb}
\usepackage{subcaption,epsf,epstopdf}
\usepackage{multirow}
\usepackage{makecell}
\newtheorem{prop}{Proposition}

\newcommand{\sqz}{\vspace{-2pt}}

\usepackage{pifont}
\usepackage{tikz}
\usepackage{calc}
\def\scalecheck{\resizebox{\widthof{\checkmark}*\ratio{\widthof{x}}{\widthof{\normalsize x}}}{!}{\checkmark}}

\begin{document}
%
\title{Adaptive Transform Domain Image Super-resolution Via Orthogonally Regularized Deep Networks}
%
%
%
%

\author{Tiantong~Guo,~\IEEEmembership{Student Member,~IEEE,}
        Hojjat~S.~Mousavi,~\IEEEmembership{Member,~IEEE,}
        and~Vishal~Monga,~\IEEEmembership{Senior~Member,~IEEE}
\IEEEcompsocitemizethanks{\IEEEcompsocthanksitem T. Guo, H. Mousavi and V. Monga are with the Department
of Electrical Engineering, The Pennsylvania State University, University Park,
PA, 16802.\protect\\
E-mail: see \url{http://signal.ee.psu.edu}}
\thanks{Manuscript received in August 2018, revised in February and April 2019.}}

%
%

\markboth{ACCEPTED TO TRANSACTIONS ON IMAGE PROCESSING, PRE-PRINT VERSION, APRIL 2019}{Guo \MakeLowercase{\textit{et al.}}: Adaptive Transform Domain Image Super-resolution Via Orthogonally Regularized Deep Networks}
\maketitle
\begin{abstract}
Deep learning methods, in particular, trained Convolutional Neural Networks (CNN) have recently been shown to produce compelling results for single image Super-Resolution (SR). Invariably, a CNN is learned to map the Low Resolution (LR) image to its corresponding High Resolution (HR) version in the spatial domain. We propose a novel network structure for learning the SR mapping function in an image transform domain, specifically the Discrete Cosine Transform (DCT). As the first contribution, we show that DCT can be integrated into the network structure as a Convolutional DCT (CDCT) layer. With the CDCT layer, we construct the DCT Deep SR (DCT-DSR) network. We further extend the DCT-DSR to allow the CDCT layer to become {\em trainable (i.e., optimizable)}. Because this layer represents an image transform, we enforce pairwise orthogonality constraints and newly formulated {\em complexity order} constraints on the individual basis functions/filters. This Orthogonally Regularized Deep SR network (ORDSR) simplifies the SR task by taking advantage of image transform domain while adapting the design of transform basis to the training image set. Experimental results show ORDSR achieves state-of-the-art SR image quality with fewer parameters than most of the deep CNN methods. A particular success of ORDSR is in overcoming the artifacts introduced by bicubic interpolation. A key burden of deep SR has been identified as the requirement of generous training LR and HR image pairs; ORSDR exhibits a much more graceful degradation as training size is reduced with significant benefits in the regime of limited training. Analysis of memory and computation requirements confirms that ORDSR can allow for a more efficient network with faster inference.
\end{abstract}



%

\section{Introduction}\label{sec:introduction}

%
%
%
%

\IEEEPARstart{I}{mage} Super-Resolution (SR) has emerged as one of the most significant ill-posed image processing and vision problems due to a variety of applications in civilian domains as well as in law enforcement \cite{park2003super}. With an increase in the number of mobile cameras and devices, enhancing resolution via a fast, memory efficient process is highly desirable.

SR problems are divided into multi-image SR \cite{farsiu2004fast, farsiu2004advances, yuan2012multiframe, li2010multi} and Single Image SR (SISR) according to the number of images required. Multi-image SR methods exploit geometric diversity in a set of LR images (of the same scene) to enhance resolution. The performance of these methods is limited by the number of LR images available and the success of geometric alignment/transformation methods that model the differences in the LR image set \cite{farsiu2004advances}.

SISR has been of more recent interest and has been addressed largely by dictionary-based and sparsity constrained learning methods and more recently via deep learning algorithms. A typical learning/example based SR approach employs two dictionaries of HR/LR images/patches \cite{mallat2010super,chang2004super,glasner2009super,yang2010image,kim2010single}. These dictionaries are often learned with sparse-coding methods to reconstruct the SR results. Many of these methods require handcrafted dictionary features which are not readily available \cite{zhang17image}. Section \ref{sec:SR} discusses these methods in detail.

Recently, deep learning methods have been shown to produce compelling state-of-the-art SR results and across a variety of different image collections \cite{Timofte_2017_CVPR_Workshops}. One of the earliest deep SR methods was SRCNN \cite{dong2014learning} and it has been extended to train multiple coupled networks \cite{tiantong16deep, bruna2015super, dong2016image,dong2016accelerating}. Other variants include \cite{wang2015self} which uses self-similar patches to explore the self-example based SR idea. Progressive \cite{lai2017deep}~and recursive networks \cite{kim2016deeply} also generate improved results with the help of diversified training data such as NTIRE\cite{Timofte_2017_CVPR_Workshops}. These spatial domain mappings were boosted by global and local bypass structures as introduced in residual learning \cite{Kim_2016_VDSR}. A key benefit of residual network structures is that they significantly reduce the training burden of the deep CNN, which is still constructed in the spatial domain.  

\textbf{Motivation}: A recent trend is deep SR but by mapping LR to HR image in the transform domain, such as the Discrete Fourier Transform (DFT) or Discrete Wavelet Transform (DWT) \cite{Timofte_2017_CVPR_Workshops,guo2017deep,Junxuan2017}. These methods show improved results by exploiting the ability of an image transform to separate coarse and fine details of an image and hence simplifying the SR task. Specifically, the DWT has been extensively explored for the SR problem in traditional model-based frameworks \cite{zhao2003wavelet,robinson2010efficient,wahed2007image,ji2009robust} and more recently also in deep networks \cite{guo2017deep}.  

We propose and develop a new adaptable transform domain deep SR method. Our starting point is the image DCT domain, in particular recognizing that the differences between a given LR-HR image pair manifest as change in high-frequency information while they typically share the same low-frequency signature (see analysis in Section \ref{sec:format}).

The \textbf{contributions} of this paper are as follows:
\begin{enumerate}
  \item We propose a novel network structure that addresses the SR problem in an image transform domain: the transform as well as its inverse are part of the network; providing an end-to-end SR mapping.
  \item We build a new  convolutional DCT (CDCT) layer integrating the DCT procedure into the Deep SR network (DCT-DSR); as a key extension we generalize the CDCT to a transform layer allowing its filters to be trainable, so that we can optimize the image transform specifically for the image SR task.
  \item We add pairwise orthogonality constraint on the newly introduced `transform layer' to allow for efficient forward and inverse transform computations. This Orthogonally Regularized Deep SR network (ORDSR) simplifies the SR task by taking advantage of image transform domain while adapting the design of transform basis to the training image set. 
  \item Inspired by the structure of DCT basis, which exhibit an increase in spatial complexity with index, we enforce a newly formulated {\em complexity order} constraint, which encourages the complexity of each learned basis to be close to its DCT counterpart.
  \item A key burden of deep SR has been identified as the requirement of generous training LR and HR image pairs; ORDSR shows a much more graceful degradation as training size is reduced with compelling improvements in the regime of limited training.
\end{enumerate}
To the best of our knowledge, ORDSR is the {\em first approach} that allows optimization of basis functions for transform domain image SR within a deep learning framework.

A preliminary version of this work has appeared as a short conference article \cite{tiantong2018ortho}. This manuscript significantly extends the 4-page conference article: First, the complexity order constraint is introduced in this work for the first time and the network structure is modified for better performance. Second, more analytical descriptions are added to explain the formulation and the training procedure of the new regularized deep network.
Third, more comprehensive experiments are reported over the short conference article. This includes detailed comparisons against state-of-the-art deep learning based SR methods and the impact of network configuration on performance, including discussions about DCT-DSR. Fourth, we extend the test image sets from Set5 \cite{bevilacqua2012low} and Set14 \cite{zeyde2010single} to additionally include BSD100 \cite{MartinFTM01} and Urban100 \cite{huang2015single}, each containing 100 test images. Fifth, a crucial new investigation is reported w.r.t varying training size(s) and ORDSR shows graceful degradation against a reduction in the number of training LR-HR pairs. Finally, an analysis of memory and computation is included to demonstrate the efficiency of ORDSR against competing alternatives.

This paper is organized as follows: Section \ref{sec:related} reviews related literature; Section \ref{sec:format} presents a new DCT domain Deep SR network (DCT-DSR) and extensions to a regularized network that allows the `transform layer' to be trainable (ORDSR).
Section \ref{sec:exp} provides experimental validation on benchmark datasets in both abundant and limited training scenarios. Section \ref{sec:conclusion} concludes the paper with thoughts for future work.

\section{Related Work}\label{sec:related}
\subsection{Single Image Super-Resolution}\label{sec:SR}

In the literature on learning-based methods for SR, sparse-coding methods have shown to be particularly effective \cite{yang2010image,yang2008image}. These techniques employ two dictionaries containing example LR and HR images/image patches. The goal is to then represent an LR image (or patch) in terms of its sparse code obtained via an LR dictionary. An HR image is obtained by using the same sparse code but applied to the dictionary of HR image patches. Several extensions of sparsity based SR have been developed including \cite{zeyde2010single,mallat2010super,kim2010single}. The focus of these methods has been to design/learn more suitable dictionaries and to find the optimal sparse representations of image patches, often by using suitable prior structure on the dictionary/sparse code \cite{mousavi2017sparsity}. 
In addition to sparse-coding based methods, self-example based methods have demonstrated success by exploring the self-similarity of the patches from the input image itself \cite{glasner2009super, yang2010exploiting, FreFat10}. 
\subsection{SR With Image Transform Domain}

In the sense of decomposing the image in terms of its different frequency components by an image transform, it is well acknowledged that the visual gaps  which need to be filled between the LR and HR images lay within the high-frequency components of the image \cite{park2003super}. Producing SR results from LR input essentially becomes a problem of recovering the high-frequency components of the image based on the LR input, whose high-frequency details are missing. Transform domain methods can enable an alternate image representation where the SR mapping may be simpler and hence learned easily and accurately. The wavelet transform has been a popular choice \cite{robinson2010efficient,zhao2003wavelet,ji2009robust,nguyen2000efficient,jiji2004single,wahed2007image,anbarjafari2010image} for traditional image SR. Recently, \cite{guo2017deep} developed a CNN network to reconstruct  wavelet coefficients of the HR image yielding significant practical improvements.

\subsection{Deep Learning for Image Super-Resolution}\label{sec:deepsr}

Other than the conventional SR methods, recently advanced computational abilities brought on by Graphic Computation Units (GPUs) have boosted research on deep CNNs for SR. These methods have quickly become the new state-of-the-art performance standard \cite{Kim_2016_VDSR, dong2016image}. Deep learning SR methods can further be divided into two classes: methods that focus on maintaining strong fidelity against ground truth HR image and those that encourage perceptually motivated and visually attractive results. A key example of the latter is Generative Adversarial Networks (GANs) \cite{goodfellow2014generative} such as \cite{johnson2016perceptual} which develops a photo-realistic styled SR method by sampling an HR image patch from an estimated distribution of the natural image patches. GAN based methods provide visually pleasing results but pay less attention to maintain pixel-value fidelity to the original HR data \cite{ledig2017photo}, which makes them unsuitable for certain practical settings. Our proposed work is consistent with a majority of the literature \cite{Timofte_2017_CVPR_Workshops} where the goal is to recover the HR image and training is based on minimizing the difference between network estimated SR and ground truth HR images.


Specifically for SR, Dong et al. introduced an SR CNN with three layers that outperformed previous sparse-coding based methods by a considerable margin and set the tone of using CNN for SR problems \cite{dong2014learning,dong2016image}. SRCNN can be viewed as a non-linear mapping function between the input LR image and the target HR image. It takes the input as a whole and uses different filters convolving with the input image to generate different feature representations which later on are convolved with following neural layers for higher level representations. It has shown promising experimental performance and great flexibility with different neural network configurations. Since then, different efforts have been made to boost CNN performance by introducing deeper structures \cite{dong2016accelerating}, utilizing residual bypasses by adding the input directly to the output of the CNN \cite{Kim_2016_VDSR}, creating different branches of networks to handle specific features \cite{Timofte_2017_CVPR_Workshops}, etc. 

Fully Connected Networks have shown a considerable improvement in SR performance by combining ideas from sparse-coding\cite{tiantong16deep}. Prior information \cite{mousavi2018deep} has shown recent promise in the deep learning framework. \cite{chen2018fsrnet} uses the face prior to help composing the human face SR images, while \cite{wang2018recovering} uses the structural feature priors to guide the network towards recovering detailed features.

Another combination of image transform domain and CNN was proposed recently \cite{Junxuan2017}. Li {\it et al.} convert an input image into its Fourier Domain and feed the DFT coefficients to the CNN. Since convolution of the filters and the image in the spatial domain is equivalent to the multiplications of the image and filters' corresponding Fourier Coefficients in Fourier Domain, they claim the operations of a CNN now becomes element-wise multiplications which speed up the training and inference of the network. Experimentally, the performance of this work is not at par with state of the art. Another limitation is the requirement of pre and post processing steps to compute the DFT/IDFT. 

Our work seeks to advance deep SR by developing an adaptable transform domain method (which we refer to as DCT-DSR). Analytically, we aim to exploit the full potential of image transforms and hence enable their explicit optimization (learning) via a new network structure and a regularized cost function (we refer to this method as ORDSR). Experimentally, our focus is on efficiency in the network: in the sense of memory, computation and the ability to succeed even in limited training regimes, which are inherent to domains such as radar and medical imaging \cite{vishal2017image,he2012learning,wu2016super ,bahrami2016reconstruction}. 
\begin{figure}\centering
\includegraphics[width=0.7\linewidth]{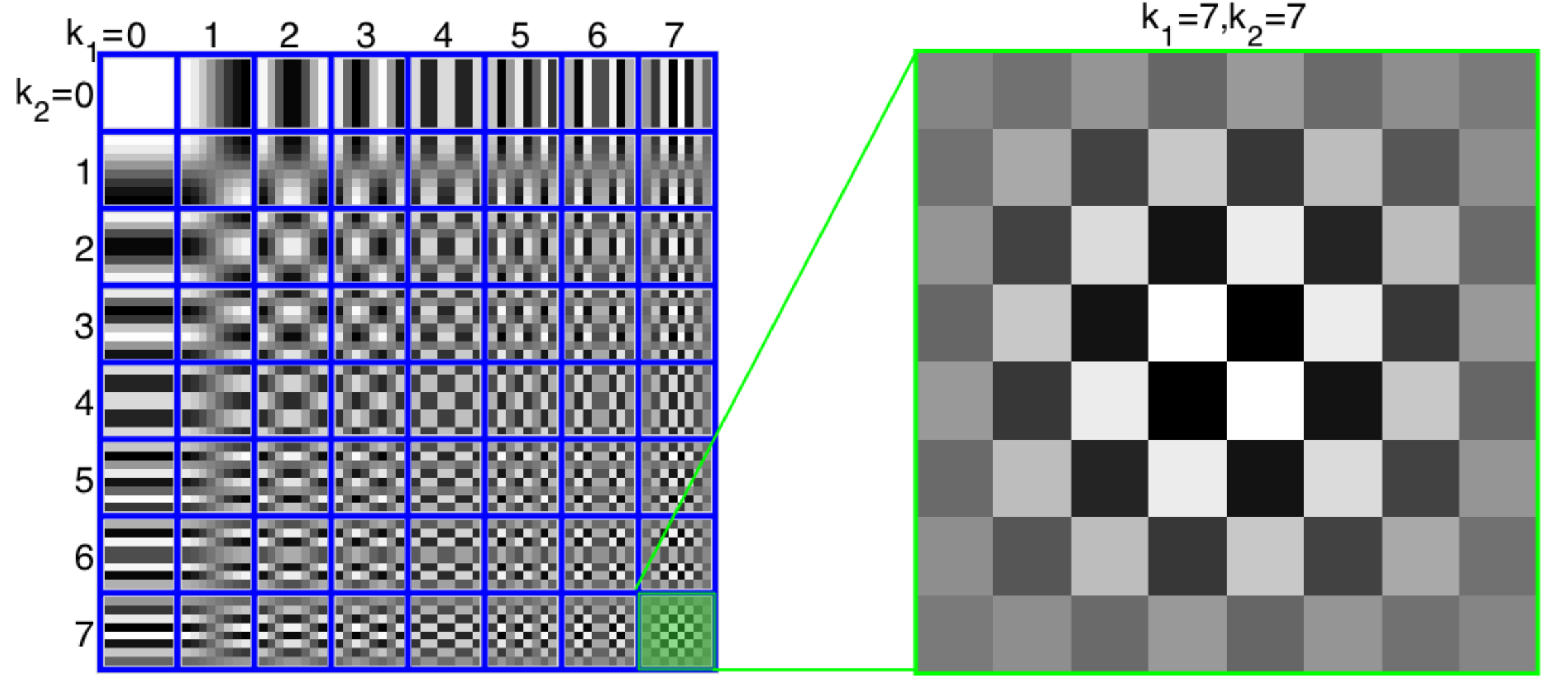}\sqz\sqz
\caption{ With $N=8$: \textit{left}: the 64 DCT basis family; \textit{right}: the last DCT basis $\mathbf{w}^\text{dct}_{7,7}$, which has size of $8\times 8$.}\label{fig:dct_basis}
\end{figure} 

\section{Orthogonally Regularized Deep SR}
\label{sec:format}

We first briefly review the DCT, IDCT and the SR problem with DCT. Then, we describe the ORDSR network structure while detailing the training and inference procedures. 

This paper uses following notations: $*$ denotes the convolution operation; $vec(\cdot)$ denotes the vectorization operation which converts the matrix into a column vector; $<\cdot,\cdot>_F$ denotes the real valued Frobenius inner product.\label{p:def}

\subsection{DCT, IDCT and Super-resolution}\label{sec:DCTIDCT}
An image $\mathbf{x}(n_1, n_2)$ of size $H\times W$ can be decomposed into $H/N\times W/N$ blocks of size $N\times N$.\footnote{We assume $H$ and $W$ are multiples of $N$ for simplicity of notation.} For the $(m,n)^{th}$ block, the DCT coefficients are computed as:
\begin{equation}\resizebox{0.9\linewidth}{!}{$
\mathbf{X}_{m,n}(k_1,k_2) = \sum\limits_{n_2=0}^{N-1}\sum\limits_{n_1=0}^{N-1} \mathbf{x}_{m,n}(n_1,n_2)\times \mathbf{w}^\text{dct}_{k_1,k_2}(n_1,n_2)$}
\end{equation}
where $k_1, k_2 \in \{0,\ldots, N-1\}$, and $\mathbf{w}^\text{dct}_{k_1, k_2}(n_1, n_2)$ is the DCT basis function, specifically DCT-II basis, defined as:
\begin{equation}\resizebox{0.9\linewidth}{!}{$
\mathbf{w}^\text{dct}_{k_1, k_2}(n_1, n_2) = C_{k_1,k_2}cos\left[\frac{\pi}{N}\left(n_1+\frac{1}{2}\right)k_1\right]cos\left[\frac{\pi}{N}\left(n_2+\frac{1}{2}\right)k_2\right]$}\label{eq:dct}
\end{equation}
where $C_{k_1, k_2} = \frac{\sqrt{1+\delta_{k_1}}\sqrt{1+\delta_{k_2}}}{N}$ and $\delta_k=1$ if $k=0$, $\delta_k=0$ otherwise. For $N=8$, there are $8\times 8$ DCT bases and each basis $\mathbf{w}^\text{dct}_{k_1,k_2}$ is of size $8\times 8$, as shown in Fig. \ref{fig:dct_basis}.

Corresponding to the DCT, the inverse DCT (IDCT) for the $(m,n)^{th}$ block is computed as:
\begin{equation}\resizebox{0.9\linewidth}{!}{$
\mathbf{x}_{m,n}(n_1,n_2) = \sum\limits_{k_2=0}^{N-1}\sum\limits_{k_1=0}^{N-1}\mathbf{X}_{m,n}(k_1,k_2)\times \mathbf{w}^\text{dct}_{k_1,k_2}(n_1,n_2)$}\label{eq:idct}
\end{equation}
Note that classical DCT is typically performed on $N\times N$ blocks of the original image \cite{gonzalez1977digital}.

\textbf{Pairwise orthogonality of Basis Functions.} The basis functions $\{\mathbf{w}^\text{dct}_{k_1,k_2}\}_{k_1,k_2 =1,1}^{N,N}\in\mathbb{R}^{N\times N}$ are pairwise orthogonal, forming an orthogonal basis family:
\begin{equation}
<\mathbf{w}^\text{dct}_{k_1,k_2},\mathbf{w}^\text{dct}_{l_1,l_2}>=\begin{cases}1,&\text{if }k_1=l_1, \text{ and }k_2=l_2\\0,&\text{Otherwise}\end{cases} \label{equ:orthoDCT}
\end{equation}
where $<\mathbf{w}^\text{dct}_{k_1,k_2},\mathbf{w}^\text{dct}_{l_1,l_2}>$ denotes the inner product of two basis functions.

We now develop a reorganization of the DCT coefficients and their computation, which we show in Section \ref{sec:netstructure} helps facilitate the implementation of DCT within a CNN.

\begin{figure}
    \centering
    \includegraphics[trim=0cm 0.2cm 0cm 0cm,clip,width=0.9\linewidth]{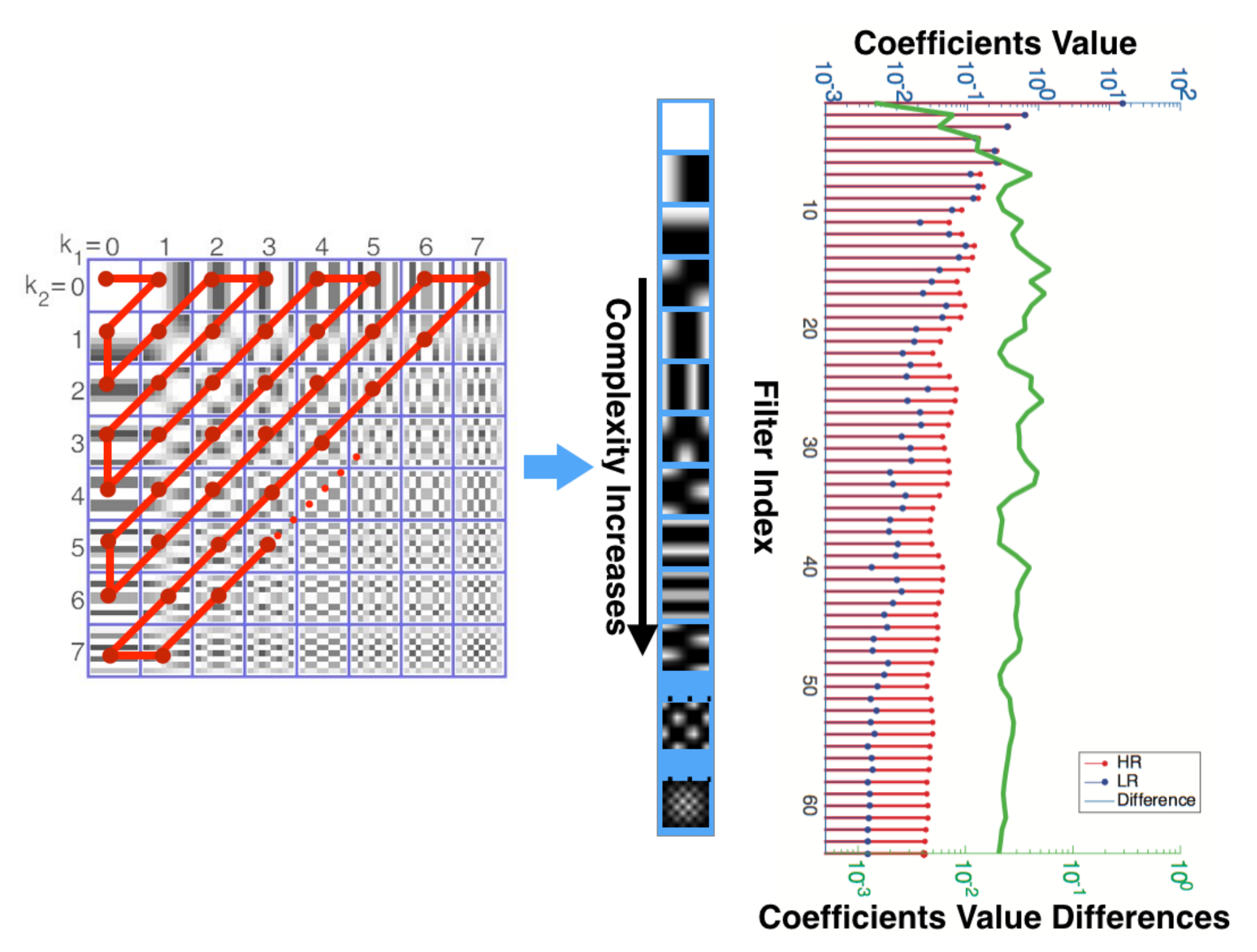}
    \caption{\textit{Left}: zig-zag reorder of the DCT basis family. \textit{Right}: average coefficient values generated by $\{\mathbf{w}^\text{dct}_i\}_{i= 1}^{64}$ of \textit{lena.bmp}.}\label{fig:zigzag}
    \end{figure}

\textbf{Zig-zag reorder.} We treat DCT basis functions as filters and reorganize them in a zig-zag order as shown in Fig. \ref{fig:zigzag}.
The zig-zag function maps $\{\mathbf{w}^\text{dct}_{k_1,k_2}\}_{k_1,k_2=1,1}^{N, N}$ to $\{\mathbf{w}^\text{dct}_i\}_{i= 1}^{N\times N}$. This reordering is similar to that used in the baseline JPEG compression procedure \cite{wallace1992jpeg}. 

\textbf{Complexity order.} Specifically, after the zig-zag reordering, as the index $i$ increases, the complexity of $\mathbf{w}^\text{dct}_i$ also increases, \textit{i.e.} the lower end (smaller $i$) of $\{\mathbf{w}^\text{dct}_i\}_{i= 1}^{N\times N}$ is corresponding to low-frequency filters, while the higher end (bigger $i$) represents the high-frequency ones.

\begin{figure*}
\centering
\includegraphics[trim=1cm 0cm 0cm 0cm,clip,width=0.8\linewidth]{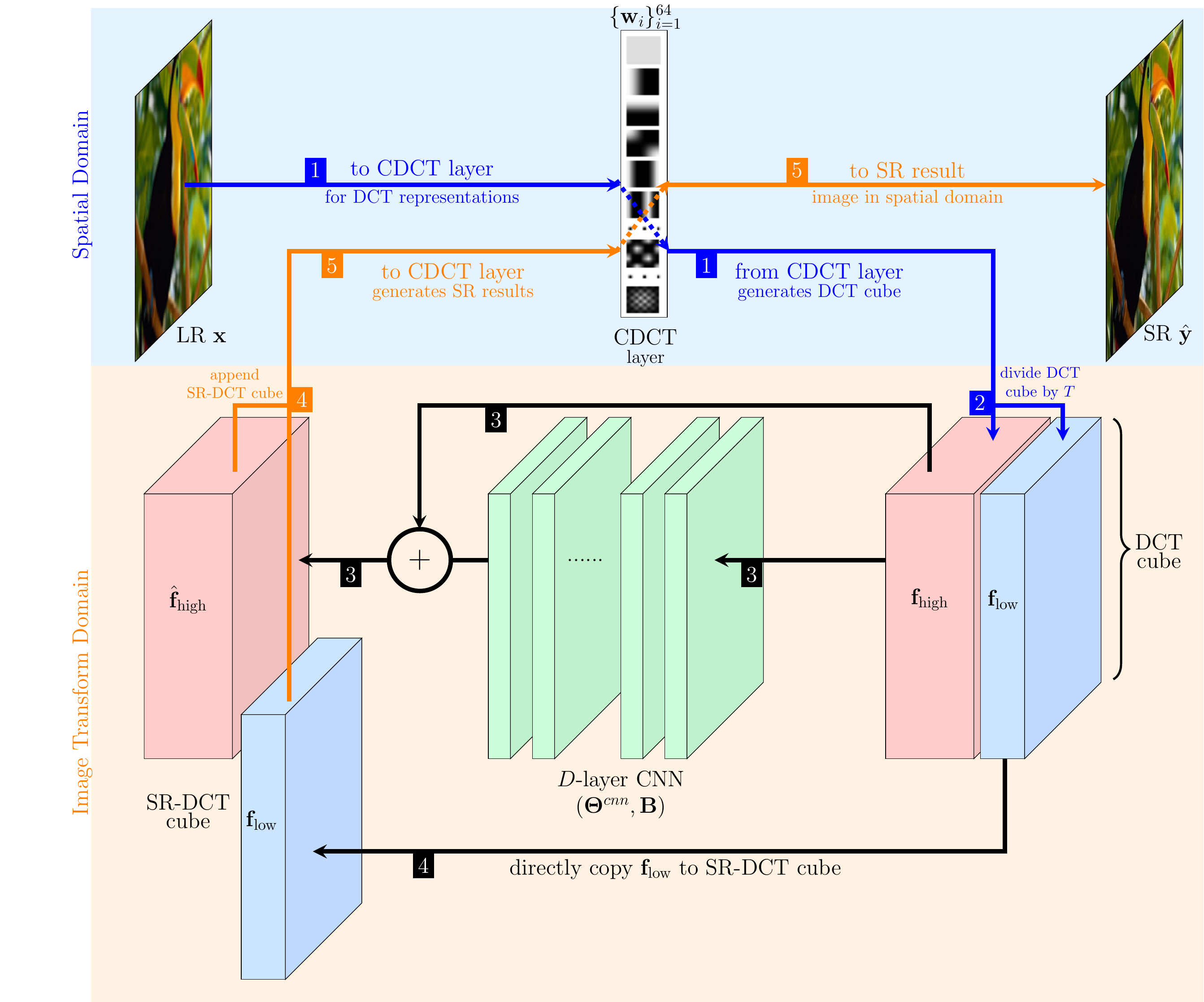}\sqz\sqz\sqz
\caption{The ORDSR network structure. Please refer to color version. Inference steps are marked with step index which is described in Section \ref{sec:test}. The CDCT layer serves two purposes: producing a DCT cube (see blue arrow) and generating SR image from an SR-DCT cube (see orange arrow).}\label{fig:ORDSR}
\end{figure*}

    Given an HR image, and its bicubic enlarged LR version\footnote{  An image is downsampled by a factor $c$ to generate the LR version, which is enlarged to its original size using bicubic interpolation.}, we can plot the average coefficient values generated by the reordered DCT filters $\{\mathbf{w}^\text{dct}_i\}_{i= 1}^{N\times N}$, as shown in Fig.\ \ref{fig:zigzag}.
    In the plot in Fig.\ \ref{fig:zigzag}, the difference between the coefficients increases with the (frequency) index. This suggests that the HR image and the LR image share the same low-frequency spectra, while they differ in high frequency content. In the DCT domain hence SR becomes the problem of recovering high-frequency DCT coefficients of the HR image from the corresponding LR ones. This insight is explicitly incorporated into the the proposed ORDSR network by focusing on reconstructing the high-frequency spectra -- see Fig.\ \ref{fig:ORDSR}.

\subsection{Network Structure}
\label{sec:netstructure}

Let us denote the bicubic enlarged LR image as $\mathbf{x}$, which is treated as preprocessing of the real input low-resolution image. Now the LR image $\mathbf{x}$ has the same size $W\times H$ as the desired HR image $\mathbf{y}$. The ORDSR network takes the $\mathbf{x}$ and produces a resolution enhanced version of $\mathbf{x}$ which is as similar as possible to $\mathbf{y}$. The network's output can be denoted as $\mathbf{\hat{y}}$. We treat the effect that the network has on the input as a nonlinear function: $F(\mathbf{x})=\mathbf{\hat{y}}$. 

The ORDSR consists of three major operations:
\begin{enumerate}
  \item {\it DCT cube representation}. The input image $\mathbf{x}$ passes through a special layer called Convolutional DCT (CDCT) layer. The outputs of the CDCT are the DCT coefficients of $\mathbf{x}$ which is referred to as the DCT cube.
  \item {\it Non-linear mapping}. The DCT cube is fed into a $D$-layer CNN for detail restoration. The CNN serves as a non-linear mapping function using the parameters learned from the training phase to restore the missing high-frequency details of the inputs. Particularly, ORDSR adopts a residual bypass structure \cite{he2016identity,Kim_2016_VDSR,guo2017deep} for faster convergence. The DCT cube is also divided into two parts which consist of low-frequency and high-frequency spectra respectively. 
  \item {\it IDCT reconstruction}. The output of the $D$-layer CNN and the low-frequency parts from the input are appended together to form a DCT cube for the SR image. The SR DCT cube is passed through the CDCT layer again (with the same filters) to reconstruct the SR image by performing transpose convolution (i.e. IDCT). 
\end{enumerate}

The overall network structure is shown in Fig. \ref{fig:ORDSR}. Next we provide a detailed description of each of the three operations mentioned above.
\subsubsection{DCT Cube Representation}\label{sec:dct}
To integrate the DCT analysis within a deep network framework, we construct a convolutional DCT (CDCT) layer. 

\textbf{Initialization.} The CDCT layer is initialized using the DCT bases $\{\mathbf{w}^\text{dct}_i\}_{i= 1}^{N\times N}$. For $N=8$, there are $64$ filters $\{\mathbf{w}_i\}_{i= 1}^{64}$ of size $8\times 8$ in the CDCT layer such that the complexity (high-frequency content) increases with the filter index. We set $N=8$ for ORDSR and from now on we take specific number 64 as the filter number of the CDCT layer .
\begin{figure}
\centering
\includegraphics[trim=0cm 0.15cm 0.15cm 0cm,clip,width=0.7\linewidth]{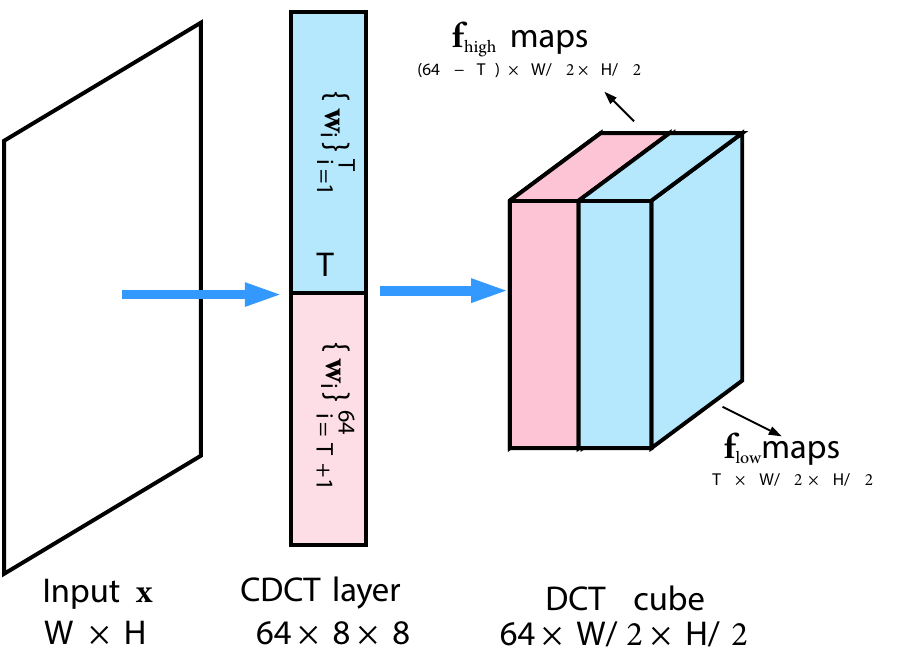}\sqz\sqz\sqz
\caption{ The CDCT layer takes an input image  and generates a DCT cube that is divided into two parts using a threshold $T$. The figure illustrates this for $N=8$, $S=2$. }\label{fig:cdct}
\end{figure}

The CDCT layer performs frequency analysis differently than traditional DCT.
Unlike classical DCT that produces $8\times 8$ block-wise DCT coefficients, the CDCT layer produces $64$ frequency maps $\{\mathbf{f}_i\}_{i= 1}^{64}$ for the whole image by convolving $\{\mathbf{w}_i\}_{i= 1}^{64}$ with the input image $\mathbf{x}$ as  in Eq. (\ref{equ:CDCTconv}) with a stride of $S$ where $*$ is the convolution operation. Note that this stride size has a significant role in the efficiency of ORDSR as analyzed in Section \ref{sec:memy}.
\begin{equation}
\mathbf{f}_i = \mathbf{w}_i * \mathbf{x} \label{equ:CDCTconv}, \forall i \in \{1,...,64\}
\end{equation}
These maps, $\{\mathbf{f}_i\}_{i= 1}^{64}$, form a cube called DCT cube. The DCT cube is essentially a reorganized version of classical block-wise DCT coefficients of the whole image as proved in the following Proposition. 
\begin{prop}
Eq \eqref{equ:CDCTconv} which performs a convolution of the input image $\mathbf x$ with the CDCT layer filters, $\mathbf{w}_i$, generates a reorganized version of the DCT coefficients of the image and is equivalent to DCT transformation.

\noindent {\it Proof:} See in Appendix \ref{sec:apx-1}. 
\end{prop}

As $i$ increases, $\mathbf{f}_i$ corresponds to higher frequency components of the whole image. Thus, we divide the DCT cube into two parts by a threshold $T$, namely low-frequency spectral maps $\mathbf{f}_\text{low} = \{\mathbf{f}_i\}^T_{i=1}$ and high-frequency spectral maps $\mathbf{f}_\text{high} = \{\mathbf{f}_i\}^{64}_{i=T +1}$, as shown in Fig. \ref{fig:cdct}. Because ORDSR uses an unconventional stride\footnote{ORDSR standard setups, details see Section \ref{sec:exp-setup}.} $S=2$, computation requirement is reduced -- see Section \ref{sec:memy} for details.

\subsubsection{Non-linear Mapping}\label{sec:CNN}

The mapping of LR to HR hi-frequency components is accomplished via a CNN consisting of $D$ convolutional layers (see Fig.\ \ref{fig:ORDSR}). Each layer has a similar operation on its input $a_l$, given by:
\begin{equation}
\mathbf{z}_l = \max(\mathbf{a}_l * \mathbf{W}_l + \mathbf{b}_l, 0) \label{equ:C}
\end{equation}
where $\mathbf{z}_l$ is the output of the $l^{th}$ layer, $\mathbf{W}_l$ and $\mathbf{b}_l$ are the weights and bias of the $l^{th}$ layer. $\mathbf{W}_l$ is a representative notation of $m_l$ filters in $l^{th}$ layer; each has a dimension of $c_l\times n_l\times n_l$. $\mathbf{b}_l$ is an $m_l$ dimensional bias vector. As is shown in Eq. (\ref{equ:C}), the convolutional layer takes the input $\mathbf{a}_l$ and applies $m_l$ convolutions on the input. This results in $m_l$ output representation maps. Then the output is processed by the ReLU operation $\max(\cdot, 0)$ \cite{nair2010rectified}.

For $l=1$, the Eq. (\ref{equ:C}) represents the processing of the input layer of the CNN, i.e.\ $\mathbf{a}_1 = \{\mathbf{f}_\text{low}, \mathbf{f}_\text{high}\}$. $\mathbf{W}_1$ is a representative notation of $m_1$ filters in layer $1$, where each of the filter has size $64\times n_1\times n_1$.

For $l=2,...,D-1$, Eq. (\ref{equ:C}) represents processing of the center layers, which takes the output from the previous layer $\mathbf{z}_{l-1}$ as its input $\mathbf{a}_l=\mathbf{z}_{l-1}$. These layers have identical structure. $\mathbf{W}_l$ is a representative notation of $m_l$ filters in $l^{th}$ layer where each of the filters has the size of $c_l\times n_l \times n_l$, which are specified in Section \ref{sec:exp}. Note that for CNNs, the number of the channels of each filter is equal to the number of filters of the previous layer, {\it i.e.} $c_l = m_{l-1}$.
\begin{equation}
\mathbf{\hat{f}}_{\text{high}} = \max(\mathbf{z}_{n-1} * \mathbf{W}_D + \mathbf{b}_D, 0)+\mathbf{f}_{\text{high}} \label{equ:C+}
\end{equation}

For $l=D$, the Eq. (\ref{equ:C+}) computes the output layer of the CNN, which produces the restored $\mathbf{\hat{f}}_\text{high}$. The output layer $\mathbf{W}_D$ is a representative notation of $(64-T)$ filters, where each of the filters has a size of $c_D\times n_D\times n_D$. The output layer generates $(64-T)$ detail maps.
The input $\mathbf{f}_\text{high}$ is added to the network output by utilizing a residual structure. Note that our choice of a residual structure is inspired by studies \cite{he2016identity,Kim_2016_VDSR,guo2017deep} which demonstrate that predicting the difference or residuals is typically a much simpler operation from an optimization standpoint. The $\mathbf{\hat{f}}_\text{high}$ serves as the final output of the $D$-layer CNN.

Collectively, let us denote the parameter sets for the CNN as $(\mathbf{\Theta}^\text{cnn}, \mathbf{B})$, where $\mathbf{\Theta}^\text{cnn}=\{\mathbf{W}_l\}_{l=1}^D$ and $\mathbf{B}=\{\mathbf{b}_l\}_{l=1}^D$. Then we denote the collective parameter sets of the ORDSR as $(\mathbf{\Theta}, \mathbf{B})$, where $\mathbf{\Theta}=\{\mathbf{\Theta}^\text{cnn},\{\mathbf{w}_i\}_{i=1}^{64}\}$, which includes the filters from the CDCT (or transform) layer.

 
\subsubsection{IDCT Reconstruction}\label{sec:idct}
Based on the restored transform coefficients $\mathbf{\hat{f}}_\text{high}$ from the $D$-layer CNN, we can generate the SR results. First, we append the $\mathbf{\hat{f}}_\text{high}$ to the $\mathbf{f}_\text{low}$ which are the low-frequency components of the input LR image as defined in Section \ref{sec:dct}. This generates an SR DCT cube  $\mathbf{f}_\text{SR} = \{\mathbf{f}_\text{low}, \mathbf{\hat{f}}_\text{high}\}$ with $64$ spectra. 

By  transpose convolving\footnote{Some literature \cite{noh2015learning,dumoulin2016guide} refer this procedure as deconvolution,  fractionally stride convolution or backward convolution in neural network setups.} the CDCT layer filters $\{\mathbf{w}_i\}_{i= 1}^{64}$ with the SR DCT cube $\mathbf{f}_\text{SR}$, the network output $\hat{\mathbf{y}}$ is generated. This procedure can be viewed as a convolution of $\mathbf{w}_i$ with a $S$ zero-padded $\mathbf{f}_i\in \mathbf{f}_\text{SR}$:

\begin{equation}
\hat{\mathbf{y}} = \sum_{i=1}^{64} \mathbf{w}_i * g_s(\mathbf{f}_i) \label{equ:CDCTdeconv}
\end{equation}
where $*$ is the convolution operation and $g_s(\cdot)$ is a $S$ zero-padding function detailed in the supplementary document \cite{techReport} (recall that $S$ is the stride used in DCT cube calculation.). Note that combined with the zero-padding function, the convolution between the $\mathbf{w}_i$ and $g_s(\mathbf{f}_i)$ can be viewed as a transposed convolutional operation between $\mathbf{w}_i$ and $\mathbf{f}_i$.\label{pg:tc}
\begin{prop} Eq \eqref{equ:CDCTdeconv} with $\mathbf{f}_\text{SR}$ as input produces a spatial image, which is equivalent to the IDCT.

\noindent {\it Proof:} See in Appendix \ref{sec:apx-2}.
\end{prop}

To summarize Section \ref{sec:dct} and \ref{sec:idct}, the CDCT layer can produce a DCT cube from an input image by performing convolution. At the same time, an image from a DCT cube can be generated by performing transpose convolution, which essentially is the IDCT. As shown in Fig. \ref{fig:ORDSR}, the CDCT layer constructs a bridge between image transform domain and the image spatial domain. 

Beyond enabling SR in the DCT domain, we show next that the basis filters of the CDCT layer can be trainable, {\it i.e. optimizable}. This opens a door towards finding customized and data-adaptive basis filters for the SR task. The optimization of CDCT/transform layer\footnote{\label{ft:DCT}After training, the filters in CDCT layer are new learned filters that can help perform a forward and inverse transform, which is indeed data-adaptive and {\em not} the DCT. 
For ease of exposition, we continue to refer to this layer as the CDCT layer and its output as the DCT cube respectively. Indeed the terms `transform layer' and `CDCT layer' are interchangeable in this paper and the context makes it clear whether the said transform is DCT or based on optimized filters/basis functions.} filters must however be constrained to yield improved results, this is detailed in the next Section.

\subsection{Desired Transform Constraints}
\label{sec:const}

While transform domain mappings can enhance SR, an image transform (viz. the proposed CDCT layer) must obey certain properties. We pose two key constraints: 
\begin{enumerate}
  \item a pairwise orthogonality constraint on filters/basis functions of the CDCT layer to guarantee reconstruction via the transpose convolution based inverse, and
  \item preservation of the complexity of the basis in terms of its order.
\end{enumerate}

\textbf{Orthogonality constraint.} The aforementioned CDCT layer can, in fact, be learned and adapted to a given training image dataset. Pairwise orthogonality constraints can be captured by a regularization term given by
\begin{equation}
\forall i\neq j, \|vec(\mathbf{w}_i)^Tvec(\mathbf{w}_j)\|^2_2\label{equ:orthoCDCT}
\end{equation}
where $i, j \in \{1,..,64\}$ and $vec(\cdot)$ is the vectorization operation which converts the matrix into a column vector.

This term is added to the network's total cost function -- see Eq. (\ref{eq:finalCostFn}). As suggested in Eq. (\ref{equ:orthoDCT}), any two distinct filter pairs in the CDCT layer should ideally have an inner product that evaluates to zero. 

\textbf{Complexity order constraint.} Because we are essentially designing a frequency domain mapping, it is desirable to preserve the order of complexity of the DCT basis. To enforce this, we introduce a new regularization term:
\begin{equation}
\|var(\mathbf{w}_t) - var(\mathbf{w}^\text{dct}_t)\|_2^2 = 0
\end{equation}
where $t\in\{1,...,64\}$, $\mathbf{w}_t$ are the filters in CDCT/transform layer and $\mathbf{w}_t^\text{dct}$ is the corresponding DCT basis function/filter (as defined in Section \ref{sec:dct}). The variance of a filter $\mathbf{w}\in \mathbb{R}^{N\times N}$ is given by Bessel's correction version \cite{probabilitybook}:
\begin{equation}
var(\mathbf{w}) = \frac{1}{N^2-1}\sum_m (\mathbf{w}^m-\frac{1}{N^2}\sum_n\mathbf{w}^n)^2
\end{equation}
where $N=8$, $\mathbf{w}^m$ and $\mathbf{w}^n$ denote an arbitrary scalar entry in filter $\mathbf{w}$. $\sum_m\mathbf{w}^m$ and $\sum_n\mathbf{w}^n$ denote the summation of all the elements inside $\mathbf{w}$. That is, we encourage the variance of the optimized filters to be close to that of their DCT counterparts. 

\subsection{Training and Inference: Regularized Optimization}
To train ORDSR we minimize a cost function that captures the functionality of the network while maintaining the properties that the CDCT layer needs to satisfy. The inference for SR procedure is then described in detailed steps.

\subsubsection{ORDSR Training}
The ORSDR network is trained by minimizing the following regularized loss function:  
\begin{equation}
\label{eq:finalCostFn}
\begin{split}
\mathbf{L}(\mathbf{\Theta},\mathbf{B}) = \underbrace{\frac{1}{2}\|F(\mathbf{x}) - \mathbf{y}\|^2_2}_{\text{MSE loss}} + \sigma\frac{1}{2}\sum_{l}^{}\sum_{m}^{m_l}\underbrace{\|\mathbf{W}_{l_m}\|^2_2}_{\text{weight decay}}\\+\gamma \frac{1}{2}\sum_{(i,j), i\neq j} \underbrace{\|vec(\mathbf{w}_i)^Tvec(\mathbf{w}_j)\|^2_2}_{\text{orthogonality constraint}}\\+\lambda\frac{1}{2}\sum_t\underbrace{\|var(\mathbf{w}_t) - var(\mathbf{w}^\text{dct}_t)\|_2^2}_{\text{complexity order constraint}}
\end{split}
\end{equation}
The cost function has four parts: Mean Square Error (MSE) loss, weight decay, orthogonality constraint and complexity order constraint. In these cost terms, MSE loss captures the similarity between the SR results $F(\mathbf{x})$ and the ground truth $\mathbf{y}$. Weight decay constraints are leveraged from the literature to prevent over-fitting \cite{goodfellow2016deep}. $\mathbf{W}_{l_m} \in \mathbf{\Theta}^\text{cnn}$ is the $m$-th weight of the CNN layer $l$ where there are $m_l$ filters in total. $\sum_l(\cdot)$ applies the weight decay term to each of the weights of the CNN and sums them together. 

Positive trade-off parameters $\gamma$ and $\lambda$ control the balance between the constraints and other cost terms.
$\sum_{(i,j)}(\cdot)$ applies the orthogonality constraint on every distinct filter pair in the CDCT/transform layer then sums them together. Similarly, $\sum_{t}(\cdot)$ applies complexity order constraint on each pair of optimized and reference (DCT) filter and sums the total.


The ORDSR is trained by using a back-propagation procedure that minimizes:
\begin{equation}
\label{eq:finalOp}
 \mathbf{\Theta},\mathbf{B} = \underset{\mathbf{\Theta},\mathbf{B}}{\arg\min} \mathbf{L}(\mathbf{\Theta},\mathbf{B})
 \end{equation}

\noindent Specifically, Eq. (\ref{eq:finalOp}) is minimized using a stochastic gradient descent method \cite{lecun1998gradient}. At iteration $t$, the CNN and the CDCT layer are updated as: $\mathbf{\Theta}^{t+1} = \mathbf{\Theta}^t - \eta \nabla_{\mathbf{\Theta}}\mathbf{L}$, where $\eta$ denotes the learning rate. 
As $\mathbf{\Theta} = \{\mathbf{\Theta}^\text{cnn}, \{\mathbf{w}_i\}_{i=1}^{64}\}$, and $\mathbf{\Theta}^\text{cnn}=\{\mathbf{W}_l\}_{l=1}^D$, the following gradients are to be computed\footnote{Note the update rules and the gradients for the bias terms are similar and are included in the supplementary document \cite{techReport}.}:
$$\frac{\partial \mathbf{L}}{\partial \mathbf{W}_l}, \frac{\partial \mathbf{L}}{\partial \mathbf{w}_i}$$
where $\mathbf{W}_l$ denotes one of the filters at $l^{th}$ layer of the CNN, representatively, and $\mathbf{w}_i$ denotes the $i^{th}$ filter in the CDCT layer. The equation for computing the gradient of an arbitrary entry within filter $\mathbf{W}_l$ in layer $l\in\{1,...,D\}$ is given by:
\begin{equation}
\resizebox{0.9\linewidth}{!}{$
\frac{\partial \mathbf{L}}{\partial \mathbf{W}^a_l} = -<(\hat{\mathbf{y}} - \mathbf{y}),\frac{\partial \mathbf{y}}{\partial \mathbf{W}^a_l}>_F + \sigma<\mathbf{W}_l,\frac{\partial \mathbf{W}_l}{\partial \mathbf{W}^a_l}>_F\label{eq:pW}$}
\end{equation}
where $\mathbf{W}^a_l$ denotes an arbitrary scalar entry within the representative filter $\mathbf{W}_l$, and $<\cdot,\cdot>_F$ denotes the real value Frobenius inner product\footnote{For two real valued matrix $\mathbf{A}$ and $\mathbf{B}$ with same dimension, $<\mathbf{A},\mathbf{B}>_F := \sum_{i,j} A_{i,j}B_{i,j}$ where $i,j$ are the indexes of the entries.}. In Eq. (\ref{eq:pW}), $\frac{\partial \mathbf{y}}{\partial \mathbf{W}^a_l}$ is computed by following the standard backpropagation rule for each layer $l$ \cite{lecun1998gradient}. For the CDCT filter $\mathbf{w}_i$, the gradient w.r.t an arbitrary scalar entry $\mathbf{w}_i^a$ is given by:
\begin{equation}\begin{split}
\frac{\partial \mathbf{L}}{\partial \mathbf{w}^a_i} = &-<(\hat{\mathbf{y}} - \mathbf{y}),\frac{\partial \mathbf{y}}{\partial \mathbf{w}^a_i}>_F \\&+ \underbrace{\gamma \sum_{(j)}\left(vec(\mathbf{w}_i)^Tvec(\mathbf{w}_j)\right)\mathbf{w}_j^a}_{\text{gradient of orthogonality constraint w.r.t $\mathbf{w}_i^a$}} \\&+ \underbrace{\lambda \frac{\partial var(\mathbf{w}_i)}{\partial \mathbf{w}_i^a}\left(var(\mathbf{w}_i) - var(\mathbf{w}^\text{dct}_i)\right)}_{\text{gradient of complexity order constraint  w.r.t $\mathbf{w}_i^a$}}\label{eq:pw}
\end{split}\end{equation}
where $\frac{\partial \mathbf{y}}{\partial \mathbf{w}^a_l}$ is computed following the standard backpropagation rule. $\frac{\partial var(\mathbf{w}_i)}{\partial \mathbf{w}_i^a}$ is the partial derivative of $var(\mathbf{w}_i)$ w.r.t $\mathbf{w}^a_i$ given by:
\begin{equation}\resizebox{\linewidth}{!}{$
  \frac{\partial var(\mathbf{w}_i)}{\partial \mathbf{w}_i^a} = \frac{2}{N^2(N^2-1)}\left[N^2\mathbf{w}_i^a-\sum_n\mathbf{w}_i^n-\sum_m\left(\mathbf{w}_i^m-\frac{1}{N^2}\sum_n\mathbf{w}_i^n\right)\right]\label{eq:pv}$}
\end{equation}
where $\mathbf{w}_i^a$ , $\mathbf{w}_i^m$, and $\mathbf{w}_i^n$ denote an arbitrary scalar entry in CDCT filter $\mathbf{w}_i$. $\sum_a\mathbf{w}_i^a$, $\sum_m\mathbf{w}_i^m$, and $\sum_n\mathbf{w}_i^n$ denote the summation of all the elements inside $\mathbf{w}_i$. Detailed notations and derivations of Eq. (\ref{eq:pW}), (\ref{eq:pw}), and (\ref{eq:pv}) can be found in the supplementary document \cite{techReport}.

The CDCT layer is initialized by the DCT filters as described in Section \ref{sec:dct} and the $D$-layer CNN is initialized using the Xavier method \cite{glorot2010understanding}. 
We use the well-known stochastic gradient descent Adam optimizer \cite{goodfellow2016deep} during the training procedure. 
We adapt gradient clip and a step gradient descent for faster training. Specific choice of numerical optimization parameters is provided in Section \ref{sec:net-setup}. 

\subsubsection{DCT-DSR Training}
Note that, without optimizing the CDCT layer filters ($\mathbf{w}_i\notin\mathbf{\Theta}$), the ORDSR is simplified to a baseline residual network performing SR in the DCT domain using a {\em fixed} CDCT layer. We call this network DCT-Deep SR (DCT-DSR). The DCT-DSR is trained by minimizing the following regularized loss function:
\begin{equation}
\resizebox{0.7\linewidth}{!}{$
\begin{split}
\mathbf{L}(\mathbf{\Theta}^\text{cnn},\mathbf{B}) = \underbrace{\frac{1}{2}\|F(\mathbf{x}) - \mathbf{y}\|^2_2}_{\text{MSE loss}} + \sigma\frac{1}{2}\sum_{l}^{}\underbrace{\|\mathbf{W}_l\|^2_2}_{\text{weight decay}}\label{eq:finalCostFn-dsr}
\end{split}
$}
\end{equation}
Experiments in Section \ref{sec:exp} demonstrates the effectiveness of using DCT transform domain for image SR. Moreover, it further emphasizes that optimizing the transform layer basis functions with CDCT  layer coefficients being learnable can significantly improve the image SR performance.


\subsubsection{Inference}\label{sec:test}
Fig. \ref{fig:ORDSR} shows the inference procedure of the ORDSR network with $N=8$.
For an input LR image $\mathbf{x}$, the goal of ORDSR is to generate its SR version $\mathbf{\hat{y}}$ as follows: 
\begin{enumerate}
\item The input LR image $\mathbf{x}$ is convolved with CDCT layer producing a DCT cube $\{\mathbf{f}_i\}_{i= 1}^{64}$ as in (\ref{equ:CDCTconv}).
\item The DCT cube of $\mathbf{x}$ is divided into $\mathbf{f}_\text{low}$ and $\mathbf{f}_\text{high}$ corresponding to low and high-frequency spectra using a threshold $T$. The exact separation process is described in Section \ref{sec:CNN};
\item A $D$-layer CNN takes the DCT cube $\{\mathbf{f}_\text{low}, \mathbf{f}_\text{high}\}$ as input and recovers the missing high-frequency information using a residual network structure, generating $\hat{\mathbf{f}}_\text{high}$.
\item The $\hat{\mathbf{f}}_\text{high}$ is appended to $\mathbf{f}_\text{low}$ forming the SR-DCT cube $\mathbf{f}_{\text{SR}}$. As the $\mathbf{f}_\text{low}$ is unchanged between $\mathbf{x}$ and its corresponding HR image, only $\mathbf{f}_\text{high}$ needs to be modified for generating $\mathbf{\hat{y}}$.
\item The SR-DCT cube $\mathbf{f}_{\text{SR}}$ is transpose convolved with the filters in the CDCT/transform layer (to perform the IDCT/inverse transform) generating $\mathbf{\hat{y}}$.
\end{enumerate}

In Step 2, the CDCT layer uses an unconventional stride $S=2$, which reduces the spatial size of the feature maps by factor of $4$. This gives the ORDSR a huge advantage in the inference speed and memory requirements compared to most state of the art methods that operate in the spatial domain.
Steps 1 and 5 are performed in the image spatial domain while Steps 2-4 are in the image transform domain where CDCT layer serves as a `bridge' between two image domains by performing DCT/IDCT.

\section{Experiments}\label{sec:exp}
\subsection{Training and Test Data}
\label{sec:exp-setup}

The widely used 291 images dataset \cite{Schulter_2015_CVPR} is used for training. The images are augmented using three methods:
\begin{enumerate}
  \item Rotating the images by $\{45^{\circ}$, $90^{\circ}$, $135^{\circ}$, $180^{\circ}$, $225^{\circ}$, $270^{\circ}$, $315^{\circ}\}$;
  \item Horizontal and vertical flip;
  \item Scaling by factors of $\{0.7, 0.8, 0.9\}$.
\end{enumerate}
The augmented images are treated as HR images and then are down-sampled by the factor of $c$. Then the down-sampled images are enlarged using bicubic interpolation by the same factor $c$ to form the LR training images.
Note that the HR image is cropped so that its width and height are multiples of $c$. All the LR/HR images are further cropped into $40\times40$ pixels sub-images with 10 pixels overlap for training. During the test phase, several standard data sets are used. Specifically, Set5 \cite{bevilacqua2012low}, Set14 \cite{zeyde2010single}, BSD100 \cite{MartinFTM01} and Urban100 \cite{huang2015single} are used to evaluate  ORDSR\footnote{\label{ft:code}Test code and networks are available at \url{http://signal.ee.psu.edu/ORDSR.html.} Detailed training schemes are included in the supplementary document.}. The metrics used for image quality assessment are PSNR, SSIM \cite{wang2004image} and Information Fidelity Criterion (IFC) \cite{sheikh2005information}.
Note that while a few published methods work with larger datasets such as DIV2K \cite{Timofte_2017_CVPR_Workshops}, ImageNet \cite{deng2009imagenet}, or MS-COCO \cite{lin2014microsoft} -- our choice of the 291 images dataset \cite{Schulter_2015_CVPR} is for consistency and fairness of comparison against a large body of competing methods that all employ this dataset.

Both training and test phases of ORDSR and DCT-DSR only utilize the luminance channel information of the input images to be consistent with literature \cite{yang2010image, dong2011image,Kim_2016_VDSR}. Chrominance channels Cb and Cr are directly enlarged by bicubic interpolation from LR images. These enlarged chrominance channels are combined with SR luminance channel to produce color SR results. Both training and test are conducted on an NVIDIA Titan X GPU (12GB) with the Tensorflow package \cite{tensorflow2015-whitepaper}. 

\subsection{Network Setup}
\label{sec:net-setup}
In the training phase, the momentum and gradient clip are set to $0.9$ and $0.5$ respectively. The learning rate is initialized to $10^{-4}$ and updated every 30 epochs with a $25\%$ decrease. The network is first initialized using the non-learnable DCT bases and random Xavier \cite{glorot2010understanding} initialization  for the CDCT layer and CNN layers, respectively. This forms the DCT-DSR network which is trained for $80$ epochs, only optimizing the CNN layers. Then the Orthogonality and Complexity Order constraints are enforced as well as including the CDCT layer in the trainable parameter set, {\it i.e.} $\mathbf{w}_i\in\mathbf{\Theta}$, forming ORDSR. ORDSR is then trained for $80$ epochs. The stride $S$ is set to $2$ to eliminate block effects as well as to reduce the memory and computational requirements (see Section \ref{sec:memy}). Unless stated otherwise, the standard configuration of ORDSR is as follows: $\gamma = 3.5$, $\lambda=0.75$, 
$D = 15$, $T=4$ (for $c=3$; for other scale factors see Section \ref{sec:T}), $m_i=64$ where $i\in\{1,...,14\}$ and $n_1=5$, $n_i=3$, where $i\in\{2,...,15\}$. All hyper-parameters are determined using cross-validation. During the training $128$ training patch pairs with the size of $40\times40$ are randomly extracted in each batch.\label{tx:train}

\begin{figure}
    \centering
    \begin{subfigure}{\linewidth}
    \centering
    \includegraphics[trim=2cm 1.6cm 2cm 1.6cm, clip, width=1\linewidth]{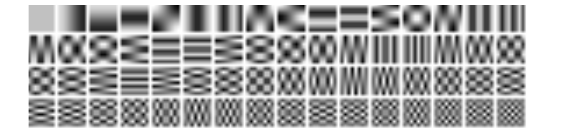}      
    \vspace{-15pt}\caption{Initialization DCT Filters, as used in DCT-DSR}\vspace{5pt}
    \end{subfigure}
    \begin{subfigure}{\linewidth}
    \centering
    \includegraphics[trim=2cm 1.6cm 2cm 1.6cm, clip, width=1\linewidth]{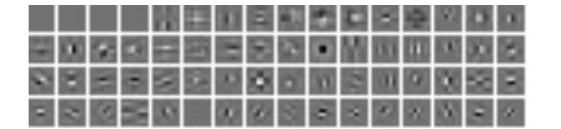}    
    \vspace{-15pt}\caption{Learned Filters, as used in ORDSR}\label{fig:e01}\vspace{5pt}
    \end{subfigure}
    \begin{subfigure}{\linewidth}
    \centering
    \includegraphics[trim=2cm 1.6cm 2cm 1.6cm, clip, width=1\linewidth]{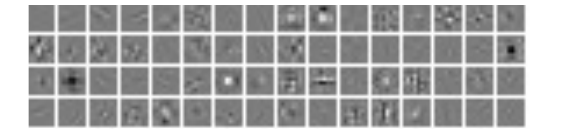} 
    \vspace{-15pt}\caption{Learned Filters, as used in ORDSR-RI}\label{fig:e01}\vspace{5pt}
    \end{subfigure}
    \caption{CDCT layer initialization and the learned filters of CDCT layer in ORDSR. (Filters are normalized and reordered for display purposes.)}\label{fig:filter}
\end{figure}

\begin{figure}[t]
\centering
\includegraphics[trim=1.2cm 0cm 2cm 0cm,clip,width=\linewidth]{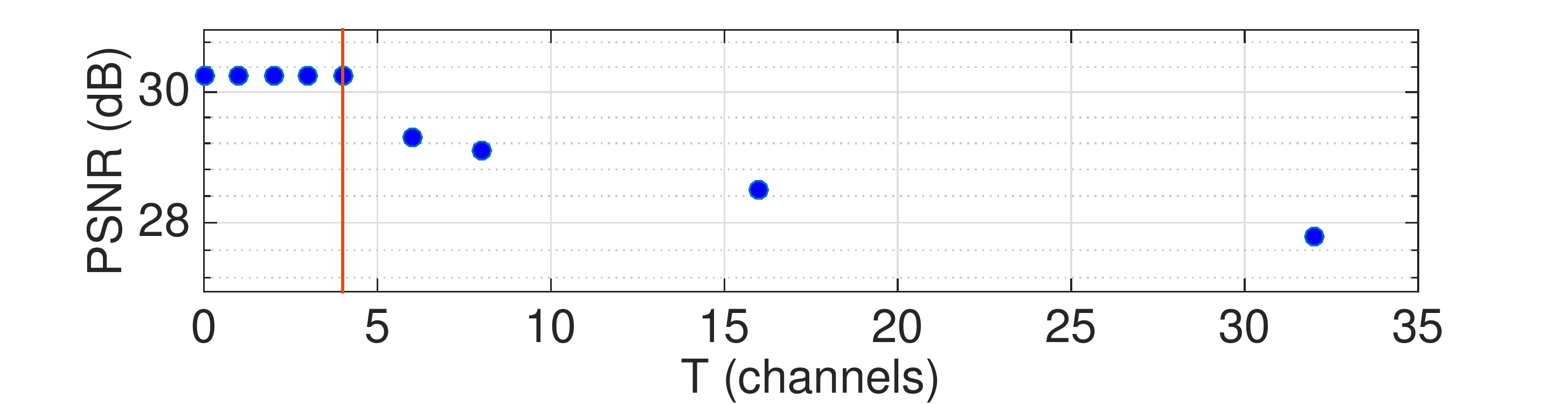}
\caption{Avg. PSNR of Set14 with scale factor 3 on different $T$. When $T<5$ , decreasing $T$ will not affect the SR results. Hence we set $T=4$ for $c=3$.}\label{fig:T}
\end{figure}

\begin{figure}
\centering
\includegraphics[trim=0cm 0cm 0cm 0cm,clip,width=\linewidth]{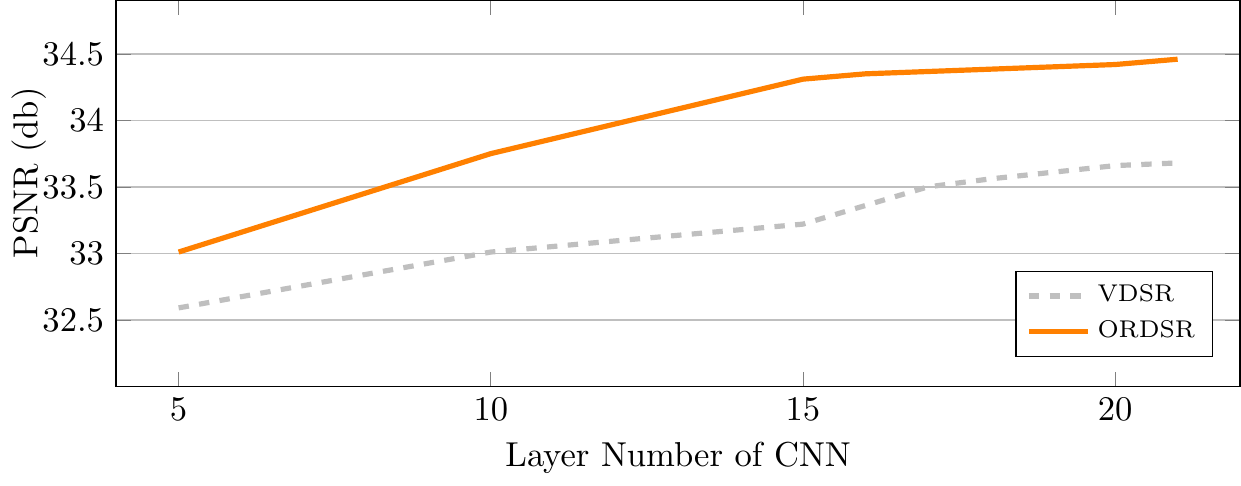}\sqz\sqz\sqz
\caption{Avg. PSNR of Set5 with scale factor 3 across different number of layers. ORDSR outperforms VDSR for varying network depths.}\label{fig:layer}
\end{figure}
\begin{figure*}[ht]
\centering
\includegraphics[trim=0cm 0cm 0cm 0cm,clip,width=\linewidth]{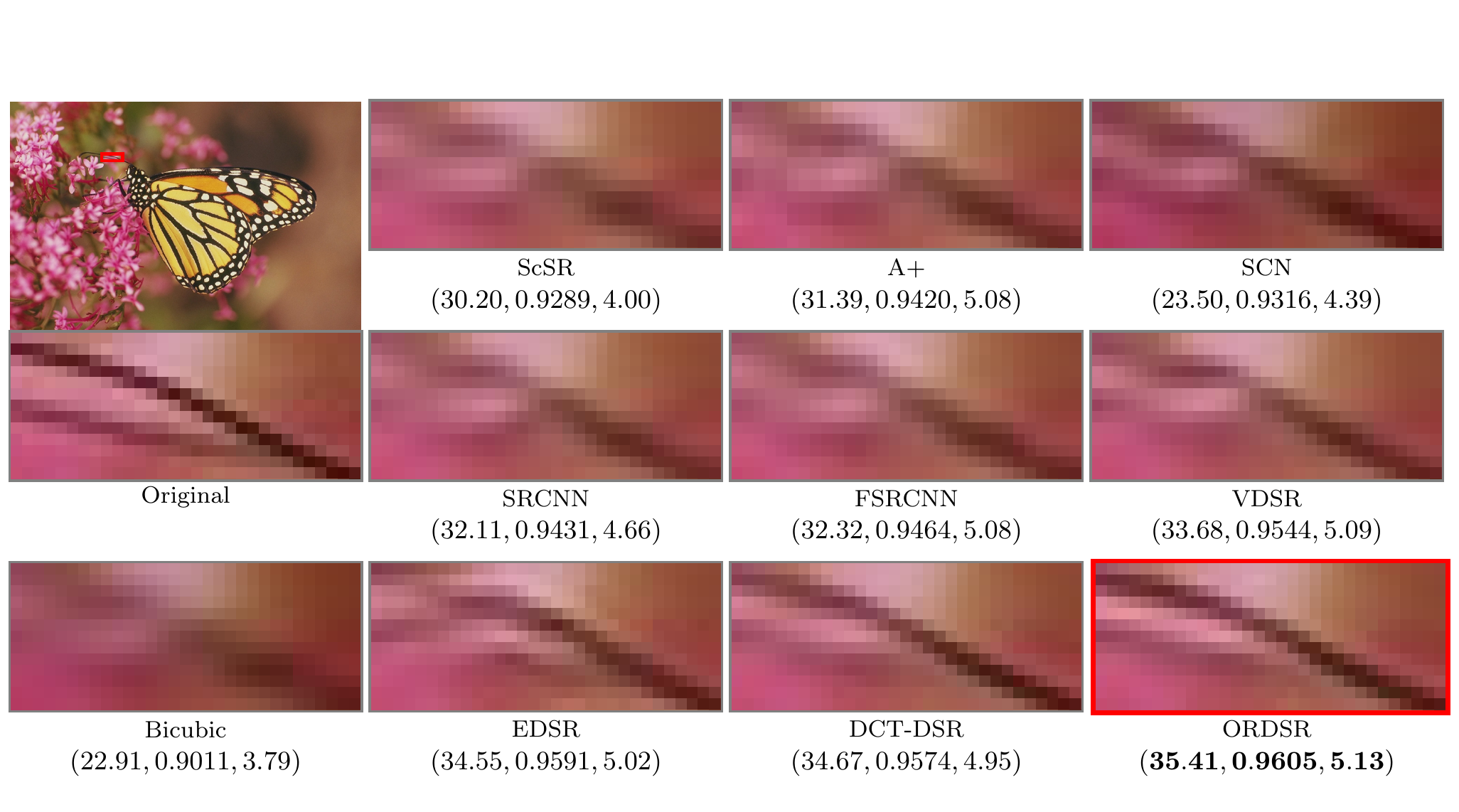}\sqz\sqz\sqz
\caption{The SR test results of \textit{monarch.bmp} for scale factor 3. The assessment metrics are shown as (PSNR, SSIM, IFC).}\label{fig:SR_but}
\end{figure*}

\begin{table}[]
\centering
\captionsetup{justification=centering}
\caption{Average PSNR, SSIM and IFC results on Set14 with scale factor 3 -- different filter size and number setups.}
\label{tb:size}\resizebox{0.95\linewidth}{!}{
\begin{tabular}{l c c c }\Xhline{3\arrayrulewidth}
 & \multicolumn{3}{c}{$m_i=32$, $i\in\{1,...14\}$} \\  \cline{2-4}
 &PSNR& SSIM& IFC\\\hline
$n_1=5$, $n_i=3$, $i\in\{2,...,15\}$&29.95&0.8322& 4.43\\
$n_i=5$, $i\in\{1,...,15\}$&29.87&0.8316&4.39\\
$n_i=3$, $i\in\{1,...,15\}$&29.92&0.8318& 4.42\\\Xhline{3\arrayrulewidth}
 & \multicolumn{3}{c}{$m_i=64$, $i\in\{1,...14\}$} \\  \cline{2-4}
 &PSNR& SSIM& IFC\\\hline
$n_1=5$, $n_i=3$, $i\in\{2,...,15\}$&\bf 30.26&\bf 0.8380&\bf 4.55\\
$n_i=5$, $i\in\{1,...,15\}$&29.96&0.8298&4.45\\
$n_i=3$, $i\in\{1,...,15\}$&30.08&0.8301&4.47\\\Xhline{3\arrayrulewidth}
\end{tabular}}
\end{table}


\subsection{Impact of ORDSR Network Parameters}
\subsubsection{Threshold T on DCT Cube}
\label{sec:T}

This threshold Separates the DCT cube into two parts as described in Section \ref{sec:DCTIDCT} and ORDSR focuses on restoring the high-frequency details $\hat{\mathbf{f}}_\text{high}$.
Fig. \ref{fig:T} shows the effect of varying $T$ on the PSNR of the SR results. A small $T$ implies a smaller fraction of DCT cube ($\mathbf{f}_\text{low}$) is directly copied to the SR-DCT cube. Setting $T=0$ means that ORDSR exploits and maps all the frequency component maps. However, Fig. \ref{fig:T} reveals, for $T<5$, decreasing the threshold does not affect SR image quality for all practical purposes. This confirms that the low-frequency spectra between LR and HR image are indeed shared. Further for scale factor of $c=2$, $T$ is found to be $5$ and for $c=4$ we select $T=3$. For smaller scale factor, more low-frequency coefficients are preserved during the downsampling, hence a bigger $T$ is suitable.\label{pg:T}

\subsubsection{Number of filters and filter size in the CNN}
In ORDSR, filters emerge from two categories: CDCT layer filters and collectively $\mathbf{\Theta}^\text{cnn}$ of the CNN. For CDCT layer, the size of the filters are predefined by the DCT basis. In this study, the DCT basis used has a filter size $8\times 8$. Same for the number of the filters, it is associated with the filter size, {\it i.e.} $8\times 8 =64$.

As has been shown in the past in many CNN based SR methods \cite{dong2011image,dong2016accelerating}, the filter size and the number of filters influences the performance of the CNN. In ORDSR, the CNN uses a residual bypass structure as in \cite{guo2017deep,Kim_2016_VDSR}. Though identical layer setups have shown effectiveness \cite{Timofte_2017_CVPR_Workshops}, some structural changes are necessary for ORDSR. From Section \ref{sec:CNN}, 
in the CNN, 
the output layer always has $m_D=(64-T)$ filters since it needs to preserve the number of spectra. Besides the fixed parameter, in Table \ref{tb:size}, we report some configurations and corresponding results of  ORDSR by changing the number and the size of filters in the CNN. 
\begin{table}[]\centering
\captionsetup{justification=centering}
\caption{Different variants of ORDSR. OC stands for Orthogonality Constraint, CC stands for Complexity order Constraint. \ding{51} means the the layer is learnable or the constraint is in place during the learning.}\label{tb:ordsr_names}
\resizebox{0.78\linewidth}{!}{\begin{tabular}{lccc}\Xhline{2\arrayrulewidth}
Notation &CDCT layer learnable& OC & CC \\\hline
ORDSR& \scalecheck&\scalecheck&\scalecheck\\
DSR-OC& \scalecheck&\scalecheck&-\\
DSR-CC& \scalecheck&-&\scalecheck\\
DSR-UC& \scalecheck&-&-\\
DCT-DSR& -&-&-\\\Xhline{2\arrayrulewidth}
\end{tabular}}
\end{table}
As is apparent from Table \ref{tb:size}, ORDSR generally benefits from an increase in the number of filters. For the filter size, Table \ref{tb:size} shows that the ORDSR benefits from the first layer having slightly bigger filters.
This indicates increasing the input receptive field can help CNN generate a better representation of the input for later use. Also, smaller filters for layers in the center of the CNN produces more favorable results. 

\subsubsection{Number of Layers in the CNN}

Going deeper is a tempting thing to do. For many a problem domain though, diminishing returns have been reported before with an increase in the number of layers \cite{goodfellow2016deep}. For ORSDR, we observe a similar trend beyond $D=15$. Figure \ref{fig:layer} reveals that ORDSR can outperform  VDSR \cite{Kim_2016_VDSR}
with $D = 15$ layers. 

One advantage of not going too deep is benefit from a memory and computational standpoint. ORDSR's merits in this regard are elaborated upon in Section \ref{sec:memy}. It is also worth noticing that, at $D=20$, the $D$-layer CNN in ORDSR and VDSR have similar structure, and ORDSR still outperforms VDSR thanks to domain inspired regularization.

\subsection{Ablation Study: Benefits of Orthogonality and Complexity Order Constraints}\label{sec:CE}

To fully investigate the effects of the proposed constraints, we now introduce different variants of our proposed method. Table \ref{tb:ordsr_names} illustrates the 5 different versions of our method and covers the cases of whether the transform/CDCT layer is learned or fixed and which constraints (if any) are active. DCT-DSR is the precursor to ORDSR where the CDCT layer is fixed with DCT filters and hence non-trainable. For DCT-DSR, the same network setup/parameters are used as ORDSR except that the $\mathbf{w}_i\notin\mathbf{\Theta}$. For DCT-DSR hence, only the $D-$layer CNN is learned in Fig.\ \ref{fig:ORDSR}.
Besides the variants in Table \ref{tb:ordsr_names}, we also add a method, where the CDCT layer before optimization is randomly initialized (as opposed to initializing with DCT basis filters) and still learned with both constraints in place. We name it as ORDSR-RI (Orthogonally Regularized Deep SR with Random Initialization). Fig. \ref{fig:filter} visualizes the filters that comprise the CDCT/transform layer in DCT-DSR, ORDSR, and ORDSR-RI. Note that the filters in ORDSR-RI are less interpretable compared to the DCT-DSR and ORDSR as the transform layer is initialized randomly.\label{pg:fig5}

Table \ref{tb:ec} shows the performance of the aforementioned networks. Performance is evaluated both in abundant (using $100\%$ of the training data) and limited training scenarios (using $10\%$ of the training data). As is shown in Table \ref{tb:ec}, in both training scenarios, ORDSR gives the best performance followed by DSR-OC. Comparing the performance between DSR-OC, -CC, and -UC, the orthogonality constraints have the strongest influence while the complexity constraints help to boost the performance in the limited training scenarios.

Comparing DCT-DSR with ORDSR results, we can observe that making the CDCT layer learnable while enforcing domain specific constraints is critical to performance improvement.  Comparing ORDSR-RI with other variants, it is clear that DCT based initialization combined with enforcing one or more constraints outperforms ORDSR-RI. Meanwhile, ORDSR-RI is better than the variant where {\em no constraints} are used when optimizing this layer (DSR-UC) -- underscoring the value of using powerful domain specific constraints.

\begin{table}
\captionsetup{justification=centering}
\centering
\caption{Performance of variants of methods based on Table \ref{tb:ordsr_names}. Table shows average (PSNR/SSIM/IFC) results on Set14 with scale factor 3.}\label{tb:ec}
\resizebox{\linewidth}{!}{
\begin{tabular}{rcc}\Xhline{2\arrayrulewidth}
 Training Data Used (\%)& $100\%$ & $10\%$\\\hline
ORDSR   & 30.26/0.8380/4.55&29.67/0.8265/4.30\\
DSR-OC& 30.18/0.8314/4.42&29.28/0.8197/4.01\\
DSR-CC& 30.02/0.8295/4.41&29.03/0.8131/3.78\\
DSR-UC& 29.51/0.8240/4.09&28.21/0.8059/3.46\\
DCT-DSR & 29.86/0.8337/4.39&29.10/0.8143/3.87\\
ORDSR-RI & 29.56/0.8280/4.17&28.76/0.8074/3.52\\\Xhline{2\arrayrulewidth}
\end{tabular}}
\end{table}

\subsection{Comparison Against State-of-the-Art SR Methods}
\label{sec:sota}

In this Section, we compare ORDSR with representative state-of-the-art methods: both sparse-coding and deep learning based methods. Our experiments are partitioned into two scenarios -- abundant and limited training. In these tests, ORDSR outperforms the state-of-the-art methods and the gains are particularly pronounced when training is limited. We also demonstrate the efficiency of the ORDSR by analyzing the network size and memory requirements. 

\begin{table*}
\centering\captionsetup{justification=centering}
\caption{PSNR comparisons over Set5, Set14, BSD100, and Urban100. 
}\sqz\sqz\sqz
\label{tb:PSNR}\resizebox{\textwidth}{!}{
\begin{tabular}{r c c c c c c c c c c c c c c c}
\Xhline{2\arrayrulewidth}
PSNR&Scale
                               & \begin{tabular}[c]{@{}c@{}}Bicubic{}{}\end{tabular}
                               & \begin{tabular}[c]{@{}c@{}}ScSR{}\\\cite{yang2010image}{}\end{tabular}
                               & \begin{tabular}[c]{@{}c@{}}A+{}\\\cite{timofte2014a+}{}\end{tabular}
                               & \begin{tabular}[c]{@{}c@{}}SelfEx{}\\\cite{huang2015single}{}\end{tabular}
                               & \begin{tabular}[c]{@{}c@{}}SCN{}\\\cite{wang2015deep}{}\end{tabular}
                               & \begin{tabular}[c]{@{}c@{}}SRCNN{}\\\cite{dong2014learning}{}\end{tabular}
                               & \begin{tabular}[c]{@{}c@{}}FSRCNN{}\\\cite{dong2016accelerating}{}\end{tabular}
                               & \begin{tabular}[c]{@{}c@{}}VDSR{}\\\cite{Kim_2016_VDSR}{}\end{tabular}
                               & \begin{tabular}[c]{@{}c@{}}DWSR{}\\\cite{guo2017deep}{}\end{tabular}
                               & \begin{tabular}[c]{@{}c@{}}RDN{}\\\cite{zhang2018residual}{}\end{tabular}
                               & \begin{tabular}[c]{@{}c@{}}EDSR{}\\\cite{lim2017enhanced}{}\end{tabular}
                               & \begin{tabular}[c]{@{}c@{}}DCT-DSR{}\\proposed{}\end{tabular}  
                               & \begin{tabular}[c]{@{}c@{}}ORDSR{}\\proposed{}\end{tabular}
                               & \begin{tabular}[c]{@{}c@{}}ORDSR+{}\\proposed{}\end{tabular}                                    \\ \hline
Set5
& \begin{tabular}[c]{@{}c@{}}x2\\ x3\\ x4\end{tabular}
& \begin{tabular}[c]{@{}c@{}}33.64\\ 30.39\\ 28.42\end{tabular}
& \begin{tabular}[c]{@{}c@{}}35.78\\ 31.34\\ 29.07\end{tabular}
& \begin{tabular}[c]{@{}c@{}}36.55\\ 32.58\\ 30.27\end{tabular} 
& \begin{tabular}[c]{@{}c@{}}36.50\\ 32.62\\ 30.32\end{tabular} 
& \begin{tabular}[c]{@{}c@{}}36.58\\ 32.61\\ 30.41\end{tabular} 
& \begin{tabular}[c]{@{}c@{}}36.66\\ 32.75\\ 30.48\end{tabular}
& \begin{tabular}[c]{@{}c@{}}36.94\\ 33.06\\ 30.55\end{tabular}
& \begin{tabular}[c]{@{}c@{}}37.52\\ 33.66\\ 31.35\end{tabular}
& \begin{tabular}[c]{@{}c@{}}37.55\\ 33.69\\ 31.98\end{tabular} 
& \begin{tabular}[c]{@{}c@{}}37.93\\ 34.19\\ 32.54\end{tabular} 
& \begin{tabular}[c]{@{}c@{}}37.62\\ 33.72\\ 32.08\end{tabular}
& \begin{tabular}[c]{@{}c@{}}37.50\\ 33.75\\ 32.05\end{tabular}
& \begin{tabular}[c]{@{}c@{}}\bf 38.08\\ \bf 34.31\\ \bf 32.62\end{tabular}
& \begin{tabular}[c]{@{}c@{}}\bf 38.12\\ \bf 34.37\\ \bf 32.65\end{tabular} \\ \hline
Set14
& \begin{tabular}[c]{@{}c@{}}x2\\ x3\\ x4\end{tabular}
& \begin{tabular}[c]{@{}c@{}}30.22\\ 27.53\\ 25.99\end{tabular}
& \begin{tabular}[c]{@{}c@{}}31.64\\ 28.19\\ 26.40\end{tabular}
& \begin{tabular}[c]{@{}c@{}}32.29\\ 29.13\\ 27.33\end{tabular}
& \begin{tabular}[c]{@{}c@{}}32.24\\ 29.16\\ 27.40\end{tabular} 
& \begin{tabular}[c]{@{}c@{}}32.35\\ 29.16\\ 27.39\end{tabular} 
& \begin{tabular}[c]{@{}c@{}}32.42\\ 29.28\\ 27.40\end{tabular}
& \begin{tabular}[c]{@{}c@{}}32.54\\ 29.37\\ 27.50\end{tabular} 
& \begin{tabular}[c]{@{}c@{}}33.02\\ 29.75\\ 28.01\end{tabular}
& \begin{tabular}[c]{@{}c@{}}33.10\\ 29.77\\ 29.98\end{tabular} 
& \begin{tabular}[c]{@{}c@{}}33.89\\ 30.13\\ 28.68\end{tabular} 
& \begin{tabular}[c]{@{}c@{}}33.56\\ 30.01\\ 27.97\end{tabular} 
& \begin{tabular}[c]{@{}c@{}}33.08\\ 29.86\\ 28.03\end{tabular} 
& \begin{tabular}[c]{@{}c@{}}\bf 34.06\\ \bf 30.26\\ \bf 28.81\end{tabular}
& \begin{tabular}[c]{@{}c@{}}\bf 34.09\\ \bf 30.30\\ \bf 28.84\end{tabular} \\ \hline
BSD100
& \begin{tabular}[c]{@{}c@{}}x2\\ x3\\x4\end{tabular}
& \begin{tabular}[c]{@{}c@{}}29.55\\ 27.21\\25.96\end{tabular}
& \begin{tabular}[c]{@{}c@{}}30.77\\ 27.72\\26.61\end{tabular}
& \begin{tabular}[c]{@{}c@{}}31.21\\ 28.18\\26.82\end{tabular}
& \begin{tabular}[c]{@{}c@{}}31.18\\ 28.30\\26.84\end{tabular}
& \begin{tabular}[c]{@{}c@{}}31.26\\ 28.58\\26.88\end{tabular} 
& \begin{tabular}[c]{@{}c@{}}31.36\\ 28.20\\26.84\end{tabular}
& \begin{tabular}[c]{@{}c@{}}31.66\\ 28.52\\26.92\end{tabular}
& \begin{tabular}[c]{@{}c@{}}31.85\\ 28.82\\27.23\end{tabular}
& \begin{tabular}[c]{@{}c@{}}31.83\\ 28.87\\27.29\end{tabular} 
& \begin{tabular}[c]{@{}c@{}}32.09\\ 29.28\\27.64\end{tabular} 
& \begin{tabular}[c]{@{}c@{}}31.97\\ 29.21\\27.32\end{tabular}
& \begin{tabular}[c]{@{}c@{}}31.72\\ 28.93\\27.35\end{tabular}
& \begin{tabular}[c]{@{}c@{}}\bf 32.24\\\bf 29.41\\\bf 27.80\end{tabular}
& \begin{tabular}[c]{@{}c@{}}\bf 32.27\\ \bf 29.43\\ \bf 27.84\end{tabular}  \\ \hline
Urban100
& \begin{tabular}[c]{@{}c@{}}x2\\ x3\\ x4\end{tabular}
& \begin{tabular}[c]{@{}c@{}}26.66\\ 24.46\\23.14\end{tabular}
& \begin{tabular}[c]{@{}c@{}}28.26\\ 25.34\\24.02\end{tabular}
& \begin{tabular}[c]{@{}c@{}}29.20\\ 26.03\\24.32\end{tabular}
& \begin{tabular}[c]{@{}c@{}}29.54\\ 25.69\\24.78\end{tabular} 
& \begin{tabular}[c]{@{}c@{}}29.52\\ 25.56\\25.13\end{tabular} 
& \begin{tabular}[c]{@{}c@{}}29.50\\ 26.24\\24.52\end{tabular}
& \begin{tabular}[c]{@{}c@{}}29.87\\ 26.35\\24.61\end{tabular}
& \begin{tabular}[c]{@{}c@{}}30.76\\ 27.14\\25.15\end{tabular}
& \begin{tabular}[c]{@{}c@{}}30.81\\ 27.07\\25.19\end{tabular} 
& \begin{tabular}[c]{@{}c@{}}31.32\\ 28.07\\25.63\end{tabular} 
& \begin{tabular}[c]{@{}c@{}}31.14\\ 27.96\\25.07\end{tabular}
& \begin{tabular}[c]{@{}c@{}}30.88\\ 27.08\\25.17\end{tabular}
& \begin{tabular}[c]{@{}c@{}}\bf31.59\\\bf28.18\\\bf 25.76\end{tabular}
& \begin{tabular}[c]{@{}c@{}}\bf31.61\\\bf28.19\\\bf 25.79\end{tabular} \\\Xhline{2\arrayrulewidth}
\end{tabular}}
\end{table*}
\begin{table*}
\centering\captionsetup{justification=centering}\captionsetup{justification=centering}
\caption{SSIM comparisons over Set5, Set14, BSD100, and Urban100. \\Higher SSIM score (max=1) corresponds to greater structural similarity.}
\label{tb:SSIM}\resizebox{\textwidth}{!}{
\begin{tabular}{r c c c c c c c c c c c c c c c c}
\Xhline{2\arrayrulewidth}
SSIM&Scale
                               & \begin{tabular}[c]{@{}c@{}}Bicubic{}{}\end{tabular}
                               & \begin{tabular}[c]{@{}c@{}}ScSR{}\\\cite{yang2010image}{}\end{tabular}
                               & \begin{tabular}[c]{@{}c@{}}A+{}\\\cite{timofte2014a+}{}\end{tabular}
                               & \begin{tabular}[c]{@{}c@{}}SelfEx{}\\\cite{huang2015single}{}\end{tabular}
                               & \begin{tabular}[c]{@{}c@{}}SCN{}\\\cite{wang2015deep}{}\end{tabular}
                               & \begin{tabular}[c]{@{}c@{}}SRCNN{}\\\cite{dong2014learning}{}\end{tabular}
                               & \begin{tabular}[c]{@{}c@{}}FSRCNN{}\\\cite{dong2016accelerating}{}\end{tabular}
                               & \begin{tabular}[c]{@{}c@{}}VDSR{}\\\cite{Kim_2016_VDSR}{}\end{tabular}
                               & \begin{tabular}[c]{@{}c@{}}DWSR{}\\\cite{guo2017deep}{}\end{tabular}
                               & \begin{tabular}[c]{@{}c@{}}RDN{}\\\cite{zhang2018residual}{}\end{tabular}
                               & \begin{tabular}[c]{@{}c@{}}EDSR{}\\\cite{lim2017enhanced}{}\end{tabular}
                               & \begin{tabular}[c]{@{}c@{}}DCT-DSR{}\\proposed{}\end{tabular}  
                               & \begin{tabular}[c]{@{}c@{}}ORDSR{}\\proposed{}\end{tabular}   & \begin{tabular}[c]{@{}c@{}}ORDSR+{}\\proposed{}\end{tabular}                            \\ \hline
Set5
& \begin{tabular}[c]{@{}c@{}}x2\\ x3\\ x4\end{tabular}
& \begin{tabular}[c]{@{}c@{}}0.9292\\ 0.8678\\ 0.8101\end{tabular}
& \begin{tabular}[c]{@{}c@{}}0.9485\\ 0.8869\\ 0.8263\end{tabular}
& \begin{tabular}[c]{@{}c@{}}0.9544\\ 0.9088\\ 0.8605\end{tabular}
& \begin{tabular}[c]{@{}c@{}}0.9538\\ 0.9092\\ 0.8640\end{tabular}
& \begin{tabular}[c]{@{}c@{}}0.9540\\ 0.9080\\ 0.8630\end{tabular} 
& \begin{tabular}[c]{@{}c@{}}0.9542\\ 0.9090\\ 0.8628\end{tabular}
& \begin{tabular}[c]{@{}c@{}}0.9558\\ 0.9140\\ 0.8657\end{tabular}
& \begin{tabular}[c]{@{}c@{}}0.9586\\ 0.9212\\ 0.8820\end{tabular}
& \begin{tabular}[c]{@{}c@{}}0.9577\\ 0.9214\\ 0.8843\end{tabular} 
& \begin{tabular}[c]{@{}c@{}}0.9590\\ 0.9213\\ 0.9015\end{tabular} 
& \begin{tabular}[c]{@{}c@{}}0.9587\\ 0.9218\\ 0.8923\end{tabular}
& \begin{tabular}[c]{@{}c@{}}0.9573\\ 0.9220\\ 0.8850\end{tabular}
& \begin{tabular}[c]{@{}c@{}}{\bf 0.9599}\\{\bf 0.9226}\\{\bf0.9060}\end{tabular} 
& \begin{tabular}[c]{@{}c@{}}{\bf 0.9602}\\{\bf 0.9229}\\{\bf0.9063}\end{tabular} \\ \hline
Set14
& \begin{tabular}[c]{@{}c@{}}x2\\ x3\\ x4\end{tabular}
& \begin{tabular}[c]{@{}c@{}}0.8683\\ 0.7737\\ 0.7023\end{tabular}
& \begin{tabular}[c]{@{}c@{}}0.8940\\ 0.7977\\ 0.7218\end{tabular}
& \begin{tabular}[c]{@{}c@{}}0.9055\\ 0.8188\\ 0.7489\end{tabular}
& \begin{tabular}[c]{@{}c@{}}0.9032\\ 0.8196\\ 0.7518\end{tabular}
& \begin{tabular}[c]{@{}c@{}}0.9050\\ 0.8180\\ 0.7510\end{tabular} 
& \begin{tabular}[c]{@{}c@{}}0.9063\\ 0.8209\\ 0.7503\end{tabular}
& \begin{tabular}[c]{@{}c@{}}0.9088\\ 0.8242\\ 0.7535\end{tabular} 
& \begin{tabular}[c]{@{}c@{}}0.9102\\ 0.8294\\ 0.7662\end{tabular}
& \begin{tabular}[c]{@{}c@{}}0.9104\\ 0.8315\\ 0.7665\end{tabular} 
& \begin{tabular}[c]{@{}c@{}}0.9138\\ 0.8349\\ 0.7792\end{tabular} 
& \begin{tabular}[c]{@{}c@{}}0.9133\\ 0.8352\\ 0.7668\end{tabular}
& \begin{tabular}[c]{@{}c@{}}0.9091\\ 0.8337\\ {0.7680}\end{tabular}
& \begin{tabular}[c]{@{}c@{}}\bf 0.9184\\\bf 0.8380\\ \bf 0.7823\end{tabular} 
& \begin{tabular}[c]{@{}c@{}}\bf 0.9187\\\bf 0.8383\\ \bf 0.7826\end{tabular} \\ \hline
BSD100
& \begin{tabular}[c]{@{}c@{}}x2\\ x3\\x4\end{tabular}
& \begin{tabular}[c]{@{}c@{}}0.8425\\ 0.7382\\0.6672\end{tabular}
& \begin{tabular}[c]{@{}c@{}}0.8744\\ 0.7647\\0.6983\end{tabular}
& \begin{tabular}[c]{@{}c@{}}0.8864\\  0.7836\\0.7087\end{tabular}
& \begin{tabular}[c]{@{}c@{}}0.8855\\ 0.7778\\0.7106\end{tabular}
& \begin{tabular}[c]{@{}c@{}}0.8850\\ 0.7910\\0.7110\end{tabular} 
& \begin{tabular}[c]{@{}c@{}}0.8879\\ 0.7863\\0.7101\end{tabular}
& \begin{tabular}[c]{@{}c@{}}0.8920\\ 0.7897\\0.7201\end{tabular} 
& \begin{tabular}[c]{@{}c@{}}0.8960\\ 0.7976\\0.7238\end{tabular}
& \begin{tabular}[c]{@{}c@{}}0.8947\\ 0.7980\\0.7243\end{tabular} 
& \begin{tabular}[c]{@{}c@{}}0.8979\\ 0.8007\\0.7316\end{tabular} 
& \begin{tabular}[c]{@{}c@{}}0.8975\\ 0.8011\\0.7276\end{tabular}
& \begin{tabular}[c]{@{}c@{}}0.8954\\ 0.7992\\{0.7285}\end{tabular}
& \begin{tabular}[c]{@{}c@{}}{\bf 0.8984}\\\bf0.8045\\{\bf 0.7363}\end{tabular} 
& \begin{tabular}[c]{@{}c@{}}{\bf 0.8986}\\\bf0.8048\\{\bf 0.7367}\end{tabular}  \\ \hline
Urban100
& \begin{tabular}[c]{@{}c@{}}x2\\ x3\\ x4\end{tabular}
& \begin{tabular}[c]{@{}c@{}}0.8408\\ 0.7349\\0.6573\end{tabular}
& \begin{tabular}[c]{@{}c@{}}0.8828\\ 0.7827\\0.7024\end{tabular}
& \begin{tabular}[c]{@{}c@{}}0.8938\\ 0.7973\\0.7186\end{tabular}
& \begin{tabular}[c]{@{}c@{}}0.8967\\ 0.7864\\0.7374\end{tabular}
& \begin{tabular}[c]{@{}c@{}}0.8970\\ 0.8016\\0.7260\end{tabular} 
& \begin{tabular}[c]{@{}c@{}}0.8946\\ 0.7989\\0.7221\end{tabular}
& \begin{tabular}[c]{@{}c@{}}0.9010\\ 0.7512\\0.7270\end{tabular} 
& \begin{tabular}[c]{@{}c@{}}0.9140\\ 0.8272\\0.7524\end{tabular}
& \begin{tabular}[c]{@{}c@{}}0.9127\\ 0.8265\\0.7591\end{tabular} 
& \begin{tabular}[c]{@{}c@{}}0.9170\\ 0.8354\\0.7755\end{tabular} 
& \begin{tabular}[c]{@{}c@{}}0.9157\\ 0.8269\\0.7582\end{tabular}
& \begin{tabular}[c]{@{}c@{}}0.9136\\ 0.8193\\{0.7608}\end{tabular}
& \begin{tabular}[c]{@{}c@{}}{\bf 0.9181}\\\bf0.8381\\{\bf 0.7787}\end{tabular} 
& \begin{tabular}[c]{@{}c@{}}{\bf 0.9183}\\\bf0.8384\\{\bf 0.7789}\end{tabular} \\\Xhline{2\arrayrulewidth}
\end{tabular}}
\end{table*}
\begin{table*}
\centering\captionsetup{justification=centering}
\caption{IFC comparisons over Set5, Set14, BSD100, and Urban100.\\ Higher IFC score indicates better alignment of natural scene statistics.}\sqz\sqz
\label{tb:IFC}\resizebox{\textwidth}{!}{
\begin{tabular}{r c c c c c c c c c c c c c c c}
\Xhline{2\arrayrulewidth}
IFC&Scale
                               & \begin{tabular}[c]{@{}c@{}}Bicubic{}{}\end{tabular}
                               & \begin{tabular}[c]{@{}c@{}}ScSR{}\\\cite{yang2010image}{}\end{tabular}
                               & \begin{tabular}[c]{@{}c@{}}A+{}\\\cite{timofte2014a+}{}\end{tabular}
                               & \begin{tabular}[c]{@{}c@{}}SelfEx{}\\\cite{huang2015single}{}\end{tabular}
                               & \begin{tabular}[c]{@{}c@{}}SCN{}\\\cite{wang2015deep}{}\end{tabular}
                               & \begin{tabular}[c]{@{}c@{}}SRCNN{}\\\cite{dong2014learning}{}\end{tabular}
                               & \begin{tabular}[c]{@{}c@{}}FSRCNN{}\\\cite{dong2016accelerating}{}\end{tabular}
                               & \begin{tabular}[c]{@{}c@{}}VDSR{}\\\cite{Kim_2016_VDSR}{}\end{tabular}
                               & \begin{tabular}[c]{@{}c@{}}DWSR{}\\\cite{guo2017deep}{}\end{tabular}
                               & \begin{tabular}[c]{@{}c@{}}RDN{}\\\cite{zhang2018residual}{}\end{tabular}
                               & \begin{tabular}[c]{@{}c@{}}EDSR{}\\\cite{lim2017enhanced}{}\end{tabular}
                               & \begin{tabular}[c]{@{}c@{}}DCT-DSR{}\\proposed{}\end{tabular}  
                               & \begin{tabular}[c]{@{}c@{}}ORDSR{}\\proposed{}\end{tabular}   & \begin{tabular}[c]{@{}c@{}}ORDSR+{}\\proposed{}\end{tabular}                              \\ \hline
Set5
& \begin{tabular}[c]{@{}c@{}}x2\\ x3\\ x4\end{tabular}
& \begin{tabular}[c]{@{}c@{}}5.72\\ 3.45\\ 2.28\end{tabular}
& \begin{tabular}[c]{@{}c@{}}6.94\\ 3.98\\ 2.57\end{tabular}
& \begin{tabular}[c]{@{}c@{}}8.48\\4.84\\3.26\end{tabular}
& \begin{tabular}[c]{@{}c@{}}7.35\\4.05\\3.12 \end{tabular}
& \begin{tabular}[c]{@{}c@{}}7.36\\4.32\\2.91\end{tabular} 
& \begin{tabular}[c]{@{}c@{}}8.05\\4.58\\ 3.01\end{tabular}
& \begin{tabular}[c]{@{}c@{}}8.06\\4.56\\2.76\end{tabular}
& \begin{tabular}[c]{@{}c@{}}8.76\\ 4.85\\ 3.36\end{tabular}
& \begin{tabular}[c]{@{}c@{}}8.69\\ 4.47\\3.31\end{tabular} 
& \begin{tabular}[c]{@{}c@{}}8.80\\ 4.74\\ 3.81\end{tabular} 
& \begin{tabular}[c]{@{}c@{}}8.77\\ 4.79\\ 3.66\end{tabular}
& \begin{tabular}[c]{@{}c@{}}8.56\\ {4.87}\\ 3.78\end{tabular}
& \begin{tabular}[c]{@{}c@{}}{\bf8.82}\\ {\bf 4.96}\\ {\bf 4.02}\end{tabular}
& \begin{tabular}[c]{@{}c@{}}{\bf8.83}\\ {\bf 4.98}\\ {\bf 4.03}\end{tabular} \\ \hline
Set14
& \begin{tabular}[c]{@{}c@{}}x2\\ x3\\ x4\end{tabular}
& \begin{tabular}[c]{@{}c@{}}5.74\\ 3.33\\ 2.18\end{tabular}
& \begin{tabular}[c]{@{}c@{}}6.83\\ 3.75\\ 2.46\end{tabular}
& \begin{tabular}[c]{@{}c@{}}7.35\\4.26\\2.94\end{tabular}
& \begin{tabular}[c]{@{}c@{}}7.05\\4.12 \\2.32 \end{tabular}
& \begin{tabular}[c]{@{}c@{}}7.08\\4.00\\2.65\end{tabular} 
& \begin{tabular}[c]{@{}c@{}}6.68\\ 3.81\\ 2.50\end{tabular}
& \begin{tabular}[c]{@{}c@{}}7.47\\4.24\\2.55\end{tabular} 
& \begin{tabular}[c]{@{}c@{}}7.53\\ 4.33\\ 2.80\end{tabular}
& \begin{tabular}[c]{@{}c@{}}7.40\\ 4.31\\2.97\end{tabular} 
& \begin{tabular}[c]{@{}c@{}}7.61\\ 4.38\\3.15\end{tabular} 
& \begin{tabular}[c]{@{}c@{}}7.58\\ 4.43\\ 3.06\end{tabular}
& \begin{tabular}[c]{@{}c@{}}7.49\\ 4.39\\ {3.11}\end{tabular}
& \begin{tabular}[c]{@{}c@{}}{\bf 7.67}\\ {\bf 4.55}\\ {\bf 3.20}\end{tabular} 
& \begin{tabular}[c]{@{}c@{}}{\bf 7.69}\\ {\bf 4.57}\\ {\bf 3.23}\end{tabular} \\ \hline
BSD100
& \begin{tabular}[c]{@{}c@{}}x2\\ x3\\x4\end{tabular}
& \begin{tabular}[c]{@{}c@{}}5.26\\ 2.98\\1.91\end{tabular}
& \begin{tabular}[c]{@{}c@{}}6.20\\ 3.14\\2.22\end{tabular}
& \begin{tabular}[c]{@{}c@{}}7.15\\ 3.23\\2.51\end{tabular}
& \begin{tabular}[c]{@{}c@{}}6.84\\ 3.80\\2.44\end{tabular} 
& \begin{tabular}[c]{@{}c@{}}6.50\\ 3.46\\2.30\end{tabular} 
& \begin{tabular}[c]{@{}c@{}}6.09\\ 3.52\\2.18\end{tabular}
& \begin{tabular}[c]{@{}c@{}}7.01\\ 3.71\\2.32\end{tabular}
& \begin{tabular}[c]{@{}c@{}}7.16\\ 3.83\\2.62\end{tabular}
& \begin{tabular}[c]{@{}c@{}}7.14 \\ 3.84\\2.57\end{tabular} 
& \begin{tabular}[c]{@{}c@{}}7.21 \\ 3.96\\2.70\end{tabular} 
& \begin{tabular}[c]{@{}c@{}}7.19\\ 3.85\\2.53\end{tabular}
& \begin{tabular}[c]{@{}c@{}}{7.22}\\ 3.87\\{2.65}\end{tabular}
& \begin{tabular}[c]{@{}c@{}}{\bf7.29}\\\bf 4.07\\{\bf 2.89}\end{tabular} 
& \begin{tabular}[c]{@{}c@{}}{\bf7.30}\\\bf 4.09\\{\bf 2.90}\end{tabular}  \\ \hline
Urban100
& \begin{tabular}[c]{@{}c@{}}x2\\ x3\\x4\end{tabular}
& \begin{tabular}[c]{@{}c@{}}5.72\\ 3.42\\2.27\end{tabular}
& \begin{tabular}[c]{@{}c@{}}6.98\\ 3.16\\2.75\end{tabular}
& \begin{tabular}[c]{@{}c@{}}8.02\\ 3.78\\3.16\end{tabular}
& \begin{tabular}[c]{@{}c@{}}7.96\\ 3.55\\3.21\end{tabular} 
& \begin{tabular}[c]{@{}c@{}}7.32\\ 3.32\\2.86\end{tabular} 
& \begin{tabular}[c]{@{}c@{}}6.66\\ 4.01\\2.63\end{tabular}
& \begin{tabular}[c]{@{}c@{}}8.13\\ 4.43\\3.02\end{tabular}
& \begin{tabular}[c]{@{}c@{}}8.27\\ 4.63\\3.40\end{tabular}
& \begin{tabular}[c]{@{}c@{}}8.30\\ 4.71\\3.39\end{tabular} 
& \begin{tabular}[c]{@{}c@{}}8.34\\ 4.92\\3.40\end{tabular} 
& \begin{tabular}[c]{@{}c@{}}8.31\\ 4.85\\{3.42}\end{tabular}
& \begin{tabular}[c]{@{}c@{}}{8.35}\\ 4.82\\3.36\end{tabular}
& \begin{tabular}[c]{@{}c@{}}{\bf 8.42}\\\bf 5.03\\{\bf 3.45}\end{tabular} 
& \begin{tabular}[c]{@{}c@{}}{\bf 8.45}\\\bf 5.06\\{\bf 3.47}\end{tabular} \\\Xhline{2\arrayrulewidth}
\end{tabular}}
\end{table*}
We select well known methods from model-based, sparse-coding and recently developed deep learning based methods: 
\begin{enumerate}
  \item ScSR \cite{yang2010image}: the most representative sparse-coding based SR method. ScSR constructs LR/HR image patch dictionaries with a shared sparse code representation of a given LR/HR image pair. 
  \item A+ \cite{timofte2014a+}: a revised version of anchored neighborhood regression SR \cite{timofte2013anchored}. 
  \item SelfEx \cite{huang2015single}: a model-based method that exploits the self-similarity within the image itself.
  \item SCN \cite{wang2015deep}: CNN based method with a sparse prior.
  \item SRCNN \cite{dong2014learning}: the most widely used CNN based SR method (the CNN consists of 3 layers). 
  \item FSRCNN \cite{dong2016accelerating}: an enhanced version of SRCNN with deeper structure and transpose convolution layer.
  \item VDSR \cite{Kim_2016_VDSR}: CNN that utilizes residual structure with a network depth of $20$ layers.
  \item DWSR \cite{guo2017deep}: CNN that utilizes residual structure in the DWT domain.
  \item RDN \cite{zhang2018residual}: residual dense network that extracts abundant local features.
  \item EDSR \cite{lim2017enhanced}: the winning entry of the NTIRE contest held at CVPR 2017 \cite{Timofte_2017_CVPR_Workshops}, which utilizes $32$ residual blocks and output branches to handle different scale factors. Each residual block contains $2$ convolutional layer and each convolutional layer has $256$ filters.
  \item DCT-DSR: as described in Section \ref{sec:CE}. 
\end{enumerate}

As is standard practice \cite{dong2016image}, to create LR test images the known HR images are down-sampled and inputs to the network are created using bicubic interpolation\footnote{Except for EDSR \cite{lim2017enhanced} which uses deconvolutional layer at end to enlarge the image to desired size.}. For fairness in comparison, comparison is focused on end-to-end deep SR models with all the deep learning models being (re)trained with the 291 image training dataset described in Section \ref{sec:exp-setup}. The $D$-layer CNN employed in ORDSR has a residual network structure, which is extendable to progressive and recursive models \cite{lai2017deep,kim2016deeply,tai2017image}. This is beyond the scope of this paper and a topic for future study.

\subsubsection{Network Size and Memory Requirements} \label{sec:memy}

Using the CDCT layer with an unconventional stride gives ORDSR a huge advantage in faster training and test with less memory requirements. A typical test image \textit{lena.bmp} of size  $512\times512$, format {\it float32} takes {\it 257KB} disk space. Feeding the test image through VDSR,  each layer produces $64$ activation maps and each feature map has same size as input image. Assuming that a prefect memory release/recycle mechanism is in-place\footnote{At any given time, only one layer's activation maps are stored.}, at any given time, VDSR requires a minimum memory of: $\text{\it 257KB} \times 64 \approx \text{\it 16MB}$.
\begin{figure}[t]
\centering
\includegraphics[width=0.98\linewidth]{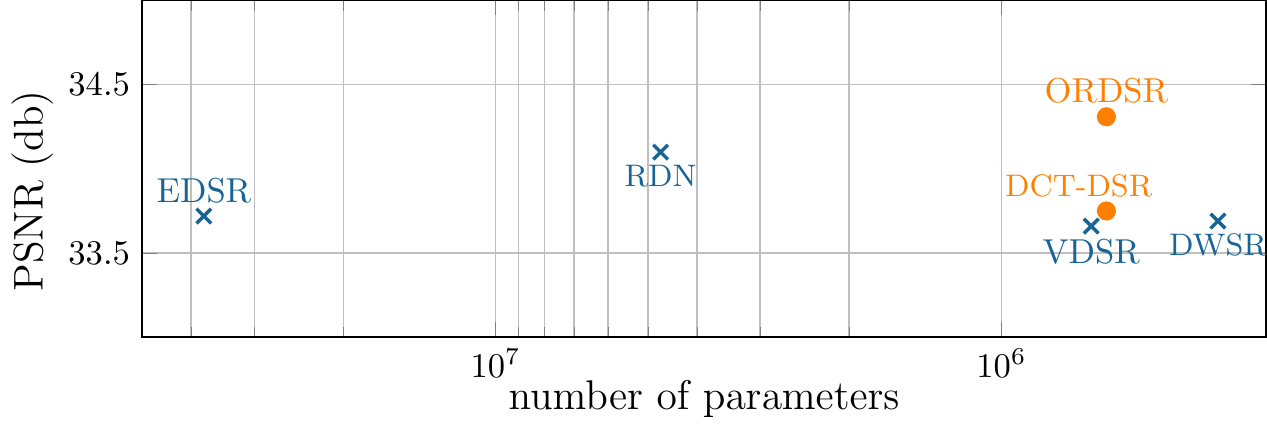}
\caption{ORDSR vs. state-of-the-art methods. PSNR on Set5 with scale factor of 3, plotted against number of parameters in the network.}
\label{fig:par}
\end{figure}
Feeding the same test image through ORDSR's CDCT layer using a stride\footnote{Other results of stride $S=3, 4, 5, 8$ can be found in the supplementary document \cite{techReport}.} of $S=2$, as shown in Fig. \ref{fig:cdct}, reduces input image width and height both by a factor of $2$. At any given time, ORDSR requires a minimum memory of $\text{\it 257KB}/2/2 \times 64 \approx \text{\it 4MB}$, which is around a quarter of VDSR. This shows ORDSR uses about four times less memory than VDSR for activation maps during the inference.  For a typical mobile camera image, which usually takes {\it 5MB} to {\it 10MB}\footnote{A standard smart-phone photo takes about {\it 8MB}.}, using ORDSR can save {\it 240MB} to {\it 480MB}. 

VDSR as reported in \cite{Kim_2016_VDSR} uses two $3\times3\times64$ and eighteen $3\times3\times64\times64$ convolutional layers. EDSR \cite{lim2017enhanced} has $32$ residual blocks where each has $2$ convolutional layers with $256$ filters in each layer. On the other hand, ORDSR in its most common realization uses: one $8\times8\times64$ CDCT layer, 
one $5\times5\times64\times64$, 
thirteen $3\times3\times64\times64$, and one $3\times3\times64\times60$ convolutional layers, which combines to produce about $44\text{\it K}$ fewer  parameters than VDSR, and about $\mathbf{95\%}$ less parameters than EDSR\footnote{For detailed computation please see supplementary document \cite{techReport}.}. During training, EDSR generates network snapshots of about $160\text{\it MB}$, while ORDSR only uses $7\text{\it MB}$ with the Tensorflow \cite{tensorflow2015-whitepaper} API. As shown in Fig. \ref{fig:par}, the ORDSR achieves better performance among all deep learning based methods while using less parameters.
\begin{figure*}
\centering
\includegraphics[trim=0cm 0cm 0cm 0cm,clip,width=\linewidth]{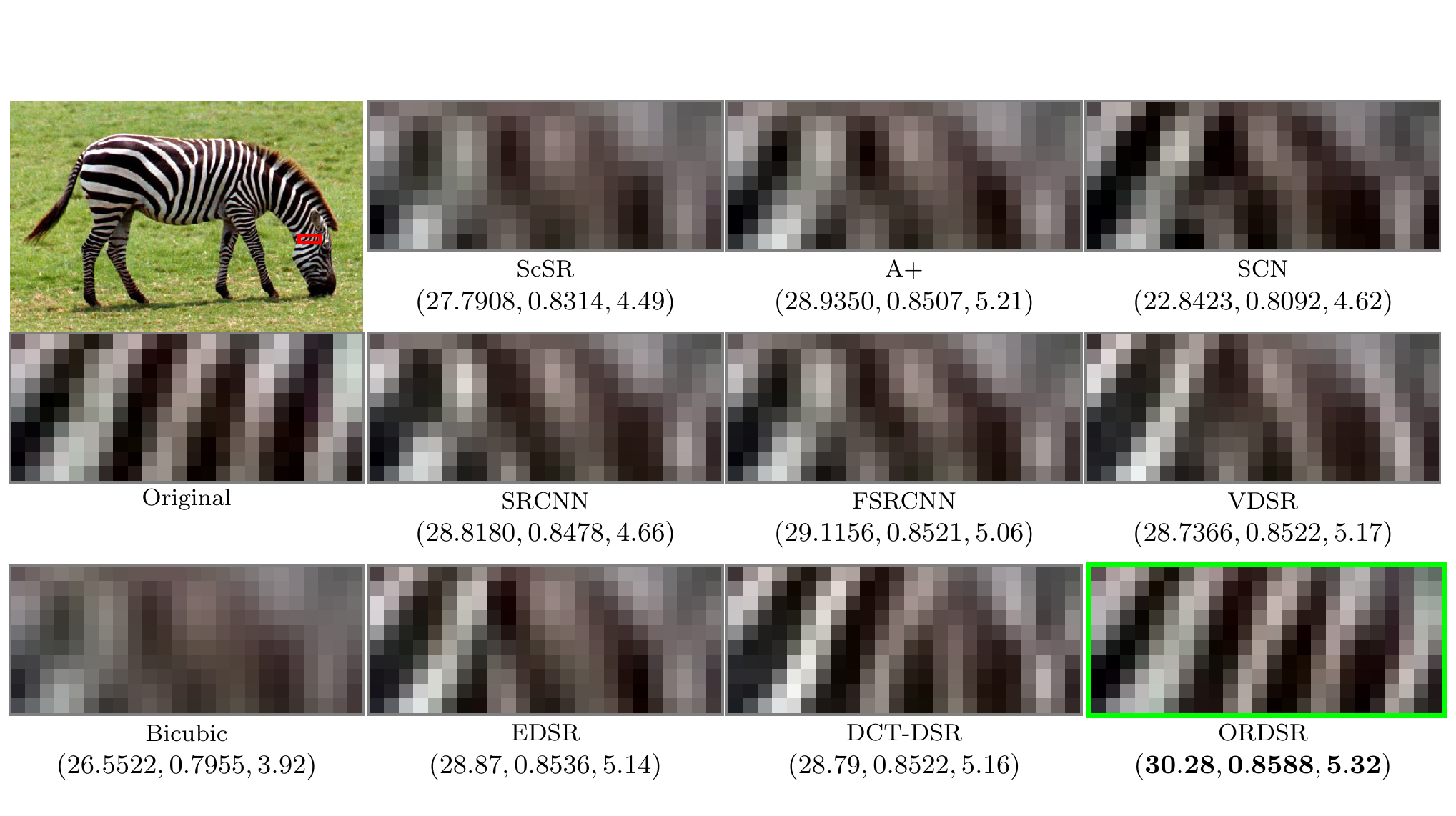}\sqz\sqz\sqz\sqz\sqz\sqz\sqz
\caption{The SR results of test image \textit{zebra.bmp} for scale factor 3. Three image quality metrics are reported along with the SR results (PSNR, SSIM, IFC). ORDSR produces best visual results, also corroborated by the quality metrics. }\label{fig:SR_zb}
\end{figure*}
\begin{figure*}
\centering
\includegraphics[trim=0cm 0cm 0cm 0cm,clip,width=\linewidth]{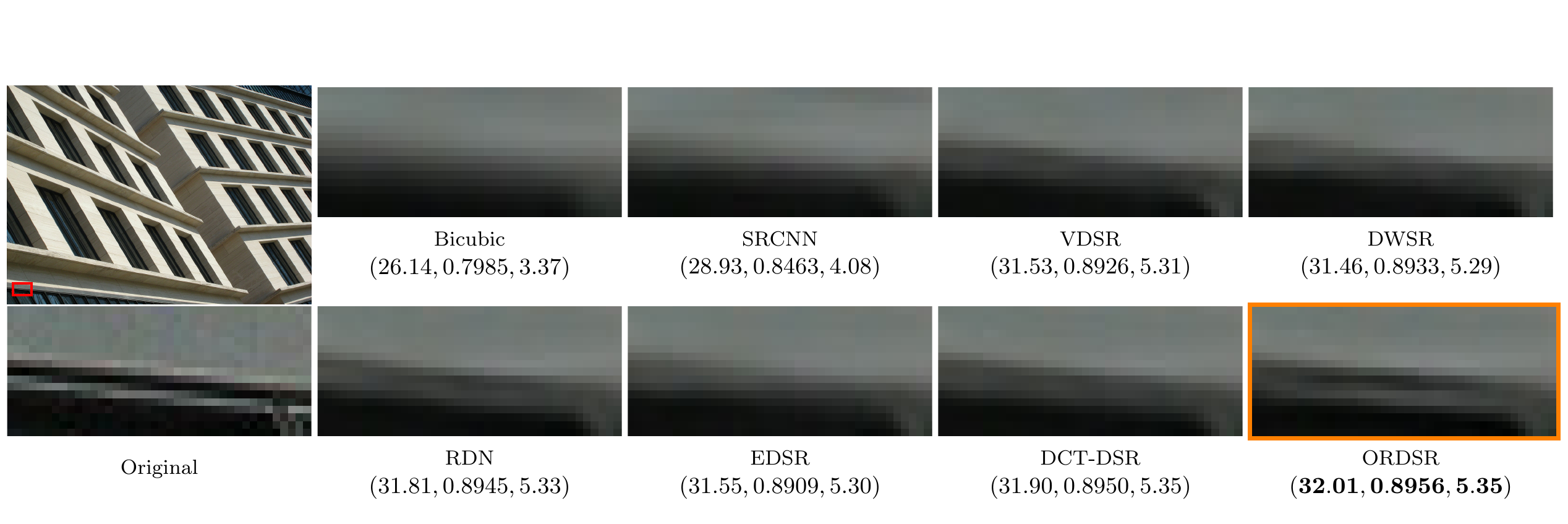}
\caption{The SR test results from Urban100 for scale factor 4. The assessment metrics are shown as (PSNR, SSIM, IFC).}\label{fig:SR_BSD}
\end{figure*}

Since ORDSR produces smaller activation maps and has a network with less parameters,  ORDSR requires less computations and trains faster than VDSR. Using the computational resources mentioned in Section \ref{sec:exp-setup}, VDSR takes $0.12 sec$ to train on one batch\footnote{Each batch is of size $128\times40\times40\times1$, for both VDSR and ORDSR.}, while ORDSR takes $0.043 sec$ per batch, which is about $2.7$ times faster.


\subsubsection{Evaluation  with Abundant Training}
\label{sec:abundant}
With $100\%$ of the training images used for training, Fig.\ \ref{fig:SR_but} and Fig.\ \ref{fig:SR_zb}
 display example test images in detail.  Note that in Fig. \ref{fig:SR_but},   deep learning based methods generate better results than sparse-coding based methods. The enlarged parts show the antennae of the monarch.  ORDSR produces more defined edges and smoother background around the antennae than competing methods.            
Tables \ref{tb:PSNR}, \ref{tb:SSIM} and \ref{tb:IFC} report the PSNR, SSIM and IFC results of ORDSR and other methods, respectively. Out of the $10$ methods reported in Tables \ref{tb:SSIM}, \ref{tb:IFC} and \ref{tb:PSNR}, VDSR, EDSR, DCT-DSR, and ORDSR produce superior results than the rest. Note that DCT-DSR can produce comparable results as EDSR and better results than VDSR by utilizing the DCT image transform domain. ORDSR further improves the performance by optimizing the transform basis. Overall, ORDSR produces best results while using $44\text{\it K}$ fewer parameters than VDSR does and $\mathbf{5\%}$ of the parameters as EDSR does.

\label{pg:self}We adopt the geometric self-ensemble strategy (similar to \cite{timofte2016seven}) to enhance the SR results. During the test, the input image $\mathbf{x}$ is flipped and rotated generating $8$ augmented versions $\mathbf{x}_i = T_i(\mathbf{x})$ where $T_i$ is one of the $8$ transformations\footnote{These transformations are: vertical flip and $\{90^{\circ},180^{\circ},270^{\circ}\}$ rotations. Combining with identity, there are $8$ versions of the input image.} including identity. Then the corresponding SR outputs $\{\hat{\mathbf{y}}_i\}_{i=1}^8$ of ORDSR are flipped and rotated back using the inverse transformation $T^{-1}(\cdot)$. The final SR result is computed as $\hat{\mathbf{y}} = \frac{1}{8}\sum_i T^{-1}(\hat{\mathbf{y}}_i)$. We mark the results using this method as ORDSR+. As is shown in Tables \ref{tb:PSNR}, \ref{tb:SSIM} and \ref{tb:IFC}, this augmentation strategy can improve the SR results mildly.
\begin{figure*}
  \centering
  \includegraphics[trim=0cm 0cm 0cm 0cm,clip,width=\linewidth]{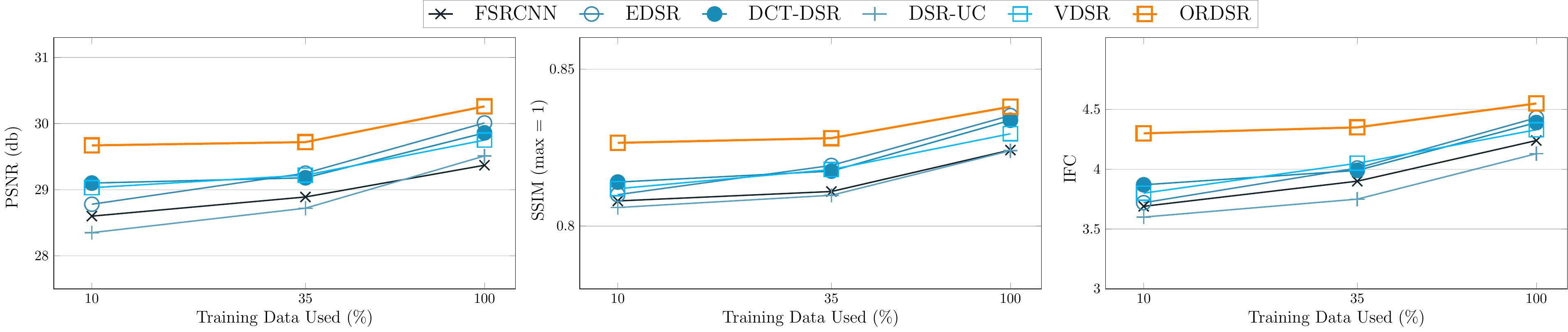}
  \begin{subfigure}{0.33\linewidth}
    \caption{PSNR results}\label{fig:lim_1}
  \end{subfigure}%
  \begin{subfigure}{0.33\linewidth}
    \caption{SSIM results}\label{fig:lim_2}
  \end{subfigure}%
  \begin{subfigure}{0.33\linewidth}
    \caption{IFC results}\label{fig:lim_3}
  \end{subfigure}
  \caption{Plots of image quality metrics (PSNR, SSIM, IFC) vs. percentage of training samples obtained for SR images for five most competitive methods -- test Set14, scale factor 3. 
  }\label{fig:lim}
\end{figure*}
\begin{figure*}
  \centering
  \includegraphics[trim=0cm 0cm 0cm 0cm,clip,width=\linewidth]{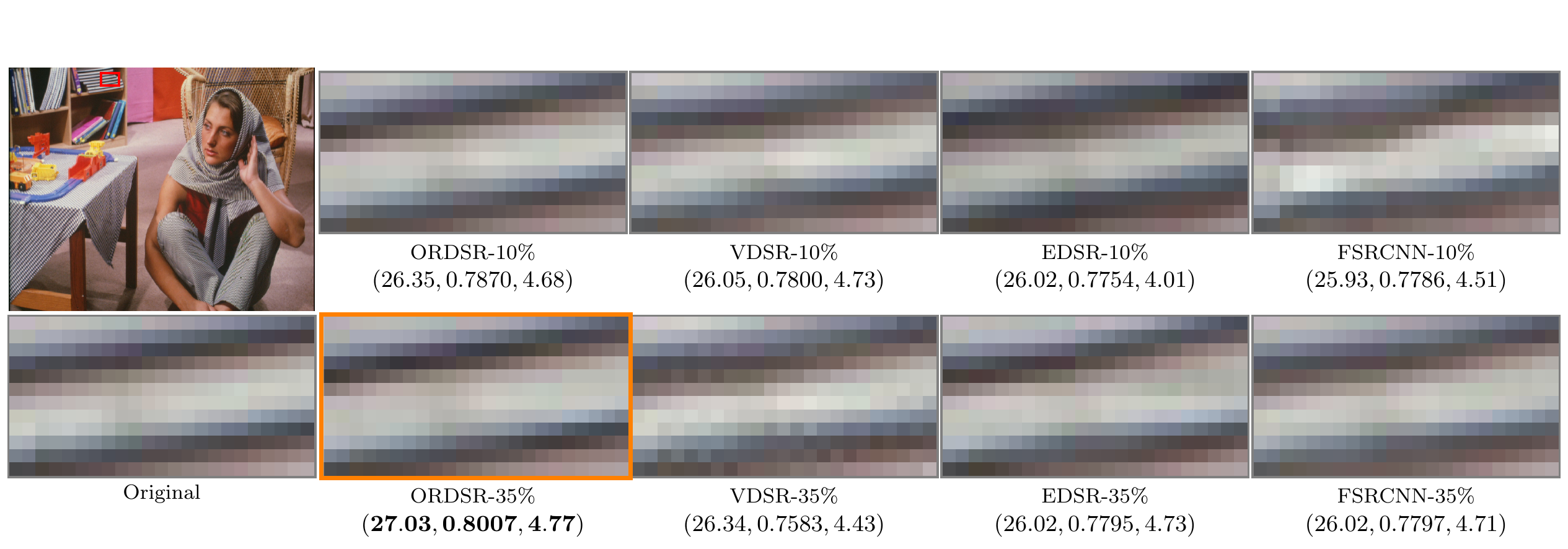}\sqz\sqz
  \caption{The SR results of test image \textit{barbara.bmp} for scale factor 3 -- results shown for $10\%$ and $35\%$, i.e.\ limited training scenarios. Three image quality metrics are reported along with the SR results (PSNR, SSIM, IFC). ORDSR produces best visual results, also corroborated by the quality metrics.}\label{fig:SR_zb_lim}
\end{figure*}

Fig. \ref{fig:SR_zb} illustrates the merits of ORDSR in overcoming artifacts introduced by bicubic interpolation and adding more details in the SR results.
In the original image of Fig. \ref{fig:SR_zb}, there is no connection between these strips. However, after downsampling and rescaling using bicubic interpolation, the artifacts are generated as `new spurious blocks' appear that connect these strips diagonally. Note that even state of the art deep learning methods are unable to overcome this. As is shown in Fig. \ref{fig:SR_zb}, for all competing methods including VDSR and EDSR, this `fake edge' is present and sometimes even enhanced. On the contrary, by virtue of operating in a carefully optimized transform domain, ORDSR exploits inter-frequency spectra information, and nearly eliminates these artifacts. In Fig. \ref{fig:SR_BSD}, the deep learning based method SR results are shown under scale factor $4$. As is shown, the ORDSR can produce more details in the SR image and has higher numerical scores. 


\subsubsection{The Limited Training Scenario}
\label{sec:limited-train}

For many real-world applications, such as medical and radar image SR \cite{vishal2017image, he2012learning, wu2016super ,bahrami2016reconstruction}, abundant training is usually not available.  We focus on two cases: $10\%$  and $35\%$ of the training image set employed in Section \ref{sec:abundant} is used. To eliminate selection bias, several random selections were made and averaged results are presented.
\begin{table*}
\centering\captionsetup{justification=centering}
\caption{Image quality metric (PSNR/SSIM/IFC) comparisons over Set5 and Set14. $10\%$ training images are used. 
}
\label{tb:lim}\resizebox{0.82\linewidth}{!}{
\begin{tabular}{r c c c c c c c c c}
\Xhline{2\arrayrulewidth}
&Scale
                               & \begin{tabular}[c]{@{}c@{}}FSRCNN{}\\\cite{dong2016accelerating}{}\end{tabular}
                               & \begin{tabular}[c]{@{}c@{}}VDSR{}\\\cite{Kim_2016_VDSR}{}\end{tabular}
                               & \begin{tabular}[c]{@{}c@{}}EDSR{}\\\cite{lim2017enhanced}{}\end{tabular}
                               & \begin{tabular}[c]{@{}c@{}}DCT-DSR{}\\ proposed{}\end{tabular}
                               & \begin{tabular}[c]{@{}c@{}}ORDSR{}\\proposed{}\end{tabular}                                   \\ \hline
Set5
&\begin{tabular}[c]{@{}c@{}}x2\\ x3\\ x4\end{tabular}
& \begin{tabular}[c]{@{}c@{}}36.18/0.9409/6.38\\ 32.23/0.9097/3.85\\ 29.87/0.8678/2.31\end{tabular}
& \begin{tabular}[c]{@{}c@{}}36.82/0.9515/7.10\\ 32.84/0.9187/4.01\\ 30.57/0.8763/2.98\end{tabular}
& \begin{tabular}[c]{@{}c@{}}36.75/0.9502/7.01\\ 32.76/0.9123/3.92\\ 30.35/0.8725/2.76\end{tabular}
& \begin{tabular}[c]{@{}c@{}}36.98/0.9533/7.23\\ 32.87/0.9192/4.03\\ 30.76/0.8772/2.97\end{tabular}
& \begin{tabular}[c]{@{}c@{}}\bf 37.24/0.9542/7.95\\\bf 33.42/0.9201/4.16\\\bf 31.14/0.8792/3.02\end{tabular} \\ \hline
Set14
&\begin{tabular}[c]{@{}c@{}}x2\\ x3\\ x4\end{tabular}
& \begin{tabular}[c]{@{}c@{}}32.29/0.8920/6.26\\ 28.60/0.8089/3.69\\ 26.82/0.7298/2.23\end{tabular}
& \begin{tabular}[c]{@{}c@{}}32.35/0.8986/6.53\\ 29.03/0.8119/3.80\\ 27.25/0.7420/2.50\end{tabular}
& \begin{tabular}[c]{@{}c@{}}32.32/0.8952/6.38\\ 28.78/0.8102/3.72\\ 27.26/0.7416/2.53\end{tabular}
& \begin{tabular}[c]{@{}c@{}}32.56/0.8979/6.62\\ 29.10/0.8143/3.87\\ 27.35/0.7482/2.62\end{tabular}
& \begin{tabular}[c]{@{}c@{}}\bf 32.98/0.9001/6.96\\\bf 29.67/0.8265/4.30\\\bf 27.89/0.7532/3.17\end{tabular} 
\\\Xhline{2\arrayrulewidth}
\end{tabular}}
\end{table*}
We focus on five methods: FSRCNN, EDSR, VDSR, DCT-DSR, and ORDSR since they are shown to be most competitive. Figs. \ref{fig:lim_1}-\ref{fig:lim_3} show the PSNR, SSIM and IFC measures plotted against percentage of training data used for these five methods. Note further that, in Section \ref{sec:abundant} we compared methods exactly as they are reported in their respective articles. Here to particularly observe and isolate the effects of training size, for fairness each network is employed with the {\em same number} of layers ($10$); where EDSR is realized with $10$ residual blocks. The plots in Figs.\ \ref{fig:lim_1}-\ref{fig:lim_3} are for a scaling factor of $3$ and on the Set14 test set. 
Two major trends emerge: a.) ORDSR offers a more graceful degradation w.r.t a decrease in the number of training samples and compelling improvements when training is limited, and b.) DCT-DSR produces results somewhat comparable to VDSR and EDSR with complementary merits in low vs. high training regimes. The competitive performance of DCT-DSR shows the value of transform domain deep SR. Finally, the gains of ORDSR over DCT-DSR, particularly when training is low ($10\%$, $35\%$ cases) emphasizes the value of regularization  in improving results.

Fig. \ref{fig:SR_zb_lim} shows the limited training scenario SR results visually for the {\em barbara.bmp} image. Compared to FRCNN, EDSR, and VDSR, ORDSR generates better visual results as well as higher numerical assessments in both the $10\%$ and $35\%$ cases. 
Table \ref{tb:lim} provides more validation for the $10\%$ training case with scale factors varying from $2$-$4$ over test sets Set5 and Set14 using $10$-layer setups for all five methods. Quite clearly, ORDSR outperforms the competition and often by a fairly significant margin.

\label{pg:lim}ORDSR does better in limited training because of two reasons: 1.) the SR mapping is simplified in the transform (for e.g. DCT) domain and hence even with limited training, the network can better approximate the non-linear mapping between LR and HR transform coefficients vs. methods that are based on spatial domain mappings, and 2.) the orthogonality and complexity regularizers play a crucial role in imparting desired structure to the transform/CDCT layer filters. 
As is readily apparent in Fig. \ref{fig:lim}, ORDSR is indeed the best, while DCT-DSR is the second best with 10 percent training because DCT-DSR also shares the two benefits mentioned above. Note also from Figs. \ref{fig:lim_1}-\ref{fig:lim_3} that DSR-UC, which optimizes the transform/CDCT layer but without {\em any} constraints, indeed does poorly in low-training while still being competitive in the 100 percent training case.
Orthogonality is indeed a crucial property for guaranteed forward and inverse transforms in our design -- DSR-UC {\em naturally leans} towards more orthogonal transform filters when driven by abundant training but this property is significantly lost when training is limited.

In Fig.\ \ref{fig:lim}, both abundant and limited training size(s) are relative to size of the test set which contains about $291$ images.

\section{Conclusion}
\label{sec:conclusion}

We develop a novel network structure to tackle the SR problem in an image transform domain. We start with DCT as the choice of image transform by proposing methods that integrate it into the network structure as a convolutional DCT (CDCT) or transform layer. We evolve the said DCT-DSR into a regularized deep network that allows for constrained optimization of basis filters that comprise the transform layer. Because orthogonality constraints are central to the transform, we call our method: Orthogonally Regularized Deep Super-Resolution (ORDSR). ORDSR is subsequently shown to outperform state of the art SR methods, particularly when training imagery (LR and HR image pairs) is limited. 

In future research, other image transform domains such as DFT and DWT can be investigated for deep SR as presented in this work. This will require explicit integration of the transform within the network structure as well as design of new specialized constraints on the transform basis to arrive at new meaningful basis that are Fourier or wavelet like.


%

\appendices
\section{Proposition 1: Sketch of the Proof}\label{sec:apx-1}
We need to show that the convolution of CDCT layer filters with the input image generates the DCT coefficients but in a zig-zag reordered form. We first define a \textbf{zig-zag mapping function} $Zig(\cdot)$, such that it maps a 2D matrix to a 1D vector following Fig. \ref{fig:zigzag}:
    $Zig(k_1,k_2)=i$ 
    where 
    $(k_1,k_2)\in[0,N-1] \footnote{Note that in the appendices, we use $[a,b]$ denotes discrete intervals.}\times[0,N-1]\rightarrow i\in[1,N\times N]$.
    Thus $Zig(k_1,k_2)=i$ and $Zig^{-1}(i)=(k_1,k_2)$\\
The proof contains two cases, one in which stride size is equal to DCT block size ($S=N$) and one in which stride size is less than DCT block size ($S<N$): we list here the key steps of the proof for each case:\\
\textbf{Case 1 $(S=N)$:} For DCT transform, we convolve the image $\mathbf{x}\in\mathbb{R}^{W\times H}$ with the DCT basis filters $\{\mathbf{w}_i\}_{i=1}^{N\times N}$. For $(m,n)\in[1,H/N]\times[1,W/N]$, $\mathbf{X}^\text{dct}_{m,n}(k_1,k_2)$ is the $(m,n)^{th}$ DCT coefficients block indexed by $(k_1,k_2)$.  For CDCT layer, convolve $x$ with $w_i$ we get: $\mathbf{X}_i^\text{cdct}:=\mathbf{x}*\mathbf{w}_i$, where $\mathbf{X}_i^\text{cdct}\in\mathbb{R}^{\frac{W}{N}\times\frac{H}{N}}$.

We then prove the proposition by mapping the DCT coefficients generated by the DCT basis to the convolutional results using the zig-zag function. For a fixed $(m, n)$,  the $\mathbf{X}^\text{cdct}_i(m,n)\in\mathbb{R}^{N^2\times1}$ can be zig-zag re-indexed into $\mathbf{X}^\text{cdct}_{m,n}(i)\in\mathbb{R}^{N\times N}$ where $i\in[1,N^2]$. Then we show for a fixed $(m,n)$, $\mathbf{X}^\text{cdct}_i(m,n) = \mathbf{X}^\text{dct}_{m,n}(k_1,k_2)$ where $i\in[0,N^2]$ and $i=Zig(k_1,k_2)=i$ for $i\in[0,N^2]$. The detailed index mapping can be found in Sec. III-A of the supplementary document \cite{techReport}.\\
\textbf{Case 2 $(S<N)$:} There is an overlapping of $(N-S)$ pixels for both DCT transform and CDCT layer. The overlapping will create reorganized DCT coefficients based on overlapped inputs. Similar to Case 1, for a fixed $(m,n)$, the key steps of the proof are unchanged (see Sec. III-B of \cite{techReport}).  

\section{Proposition 2: Sketch of the Proof}\label{sec:apx-2}
Similar to Appendix \ref{sec:apx-1}, we need to show CDCT layer and IDCT transform result in the same spatial image.
We define a \textbf{zero-padding function},  $g_s(\cdot)$: For a given location $(p,q)\in[1,W]\times[1,H]$: 
    \begin{equation}\resizebox{\linewidth}{!}{$
        g_s(\mathbf{X}_i):=\bar{\mathbf{X}}_i(p,q) =\begin{cases}\frac{1}{(N/S)^2}\mathbf{X}_i(k,l),&\text{if }p= k\times S\text{ and }q=l\times S\\0,&\text{Otherwise}\end{cases},\label{xhat_N}$}
    \end{equation}
    where $k\in[1,\frac{W}{S}], l\in[1,\frac{H}{S}]$. Note that with $S=N$, the term $\frac{1}{(N/S)^2} = 1$ means that the $g_s(\cdot)$ keeps the $\mathbf{X}_i$ value as given and does not apply reweighing. This is a different case for $S< N$. We denote $g_s(\mathbf{X}_i):=\bar{\mathbf{X}}_i\in\mathbb{R}^{W\times H}$ which is a zero padded version of $\mathbf{X}_i$. Note that the transpose convolution between $\mathbf{w}_i$ and the $\mathbf{X}_i$ is the convolution between the $\mathbf{w}_i$ and the $g_s(\mathbf{X}_i)$.\\
\textbf{Case 1 $(S=N)$:} for a fixed block $(m,n)$, $k_1,k_2,n_1,n_2 \in [1,N]$, IDCT by DCT basis, $\mathbf{x}^\text{dct}_{m,n}(n_1,n_2)$ is the recovered image given as in Eq. \eqref{eq:idct}.
On the other hand, IDCT using the CDCT layer is computed as:
    \begin{equation}\resizebox{\linewidth}{!}{$
        \mathbf{x}^\text{cdct}_{m,n}(n_1,n_2)=\sum_{i=1}^{N \times N}\sum_{p=0}^{N-1}\sum_{q=0}^{N-1}\bar{\mathbf{X}}_i(m\times N + n_1-p,n\times N + n_2-q)\times \mathbf{w}_{i}(p,q)\label{idct_N}$}
    \end{equation}
    where $\bar{\mathbf{X}}_i(\cdot,\cdot)\neq 0$, while $n_1-p$ and $n_2-q$ are  multiples of $N$ (based on the definition of $\bar{\mathbf{X}}$ from Eq. \eqref{xhat_N}). We then use $Zig^{-1}$ to reorder the indices to show that $\mathbf{x}^\text{dct}_{m,n}(n_1,n_2) = \mathbf{x}^\text{cdct}_{m,n}(n_1,n_2)$ in Sec. IV-A of \cite{techReport}.\\
\textbf{Case 2 $(S<N)$:} with overlapping of $N-S$ pixels, the aforementioned padding function remains the same but with $k\in[1,\frac{W}{S}], l\in[1,\frac{H}{S}]$. For a fixed block $(m,n)$, $k_1$, $k_2$, $n_1$, and $n_2 \in [1,N]$\\
    IDCT by CDCT layer now is a weighted version of \eqref{idct_N} with $\frac{1}{(N/S)^2}$. When constructing the final output, at any given block $(m,n)$ for location $(n_1, n_2)$, 
    \begin{equation*}\resizebox{\linewidth}{!}{$
    \mathbf{x}^\text{cdct}_{m,n}(n_1,n_2) = \sum_{k=-N/2S}^{N/2S}\sum_{l=-N/2S}^{N/2S} \bar{\mathbf{x}}^\text{cdct}_{m-k,n-l}(n_1-k\times S, n_2-l\times S)\label{idct_S}$}
\end{equation*}
    with the repeated elements in the summation as detailed in Sec. IV-A of \cite{techReport}, the overlapping effect is canceled by
\begin{equation*}\resizebox{\linewidth}{!}{$
\mathbf{x}^\text{cdct}_{m,n}(n_1,n_2) = (N/S)^2\sum_i^{N\times N} \frac{1}{(N/S)^2} \mathbf{X}_i(m,n)\times\mathbf{w}_i(n_1,n_2)$}
\end{equation*}
We then use $Zig^{-1}$ to reorder the indices to show that $\mathbf{x}^\text{dct}_{m,n}(n_1,n_2) = \mathbf{x}^\text{cdct}_{m,n}(n_1,n_2)$ in Sec. IV-B of \cite{techReport}.
\vspace{-5pt}
\bibliographystyle{IEEEbib_ini}
\bibliography{refs}\vfill

\begin{thebibliography}{10}

\bibitem{park2003super}
S.~C. Park, M.~K. Park, and M.~G. Kang,
\newblock ``Super-resolution image reconstruction: a technical overview,''
\newblock {\em IEEE Signal Processing Magazine}, vol. 20, no. 3, pp. 21--36,
  2003.

\bibitem{farsiu2004fast}
S. Farsiu, M.~D. Robinson, M. Elad, and P. Milanfar,
\newblock ``Fast and robust multiframe super resolution,''
\newblock {\em IEEE Trans. on Image Processing}, vol. 13, no. 10, pp.
  1327--1344, 2004.

\bibitem{farsiu2004advances}
S. Farsiu, D. Robinson, M. Elad, and P. Milanfar,
\newblock ``Advances and challenges in super-resolution,''
\newblock {\em Int. Journal of Imaging Systems and Technology}, vol. 14, no. 2,
  pp. 47--57, 2004.

\bibitem{yuan2012multiframe}
Q. Yuan, L. Zhang, and H. Shen,
\newblock ``Multiframe super-resolution employing a spatially weighted total
  variation model,''
\newblock {\em IEEE Trans. on Circuits and Systems for Video Technology}, vol.
  22, no. 3, pp. 379--392, 2012.

\bibitem{li2010multi}
X. Li, Y. Hu, X. Gao, D. Tao, and B. Ning,
\newblock ``A multi-frame image super-resolution method,''
\newblock {\em Signal Processing}, vol. 90, no. 2, pp. 405--414, 2010.

\bibitem{mallat2010super}
S. Mallat and G. Yu,
\newblock ``Super-resolution with sparse mixing estimators,''
\newblock {\em IEEE Trans. on Image Processing}, vol. 19, no. 11, pp.
  2889--2900, 2010.

\bibitem{chang2004super}
H. Chang, D.-Y. Yeung, and Y. Xiong,
\newblock ``Super-resolution through neighbor embedding,''
\newblock in {\em IEEE Conf. on Computer Vision and Pattern Recognition}, 2004,
  vol.~1, pp. 1275–--1282.

\bibitem{glasner2009super}
D. Glasner, S. Bagon, and M. Irani,
\newblock ``Super-resolution from a single image,''
\newblock in {\em IEEE Int. Conf. on Computer Vision}, 2009, pp. 349--356.

\bibitem{yang2010image}
J. Yang, J. Wright, T.~S. Huang, and Y. Ma,
\newblock ``Image super-resolution via sparse representation,''
\newblock {\em IEEE Trans. on Image Processing}, vol. 19, no. 11, pp.
  2861--2873, 2010.

\bibitem{kim2010single}
K.~I. Kim and Y. Kwon,
\newblock ``Single-image super-resolution using sparse regression and natural
  image prior,''
\newblock {\em IEEE Trans. on Pattern Analysis and Machine Intelligence}, vol.
  32, no. 6, pp. 1127--1133, 2010.

\bibitem{zhang17image}
L. Zhang and W. Zuo,
\newblock ``Image restoration: From sparse and low-rank priors to deep priors,
  lecture notes,''
\newblock {\em IEEE Signal Processing Magazine}, vol. 34, no. 5, pp. 172--179,
  2017.

\bibitem{Timofte_2017_CVPR_Workshops}
R. Timofte, E. Agustsson, L. Van~Gool, M.-H. Yang, L. Zhang, et~al.,
\newblock ``Ntire 2017 challenge on single image super-resolution: Methods and
  results,''
\newblock in {\em IEEE Conf. on Computer Vision and Pattern Recognition
  Workshops}, July 2017, pp. 1110--1121.

\bibitem{dong2014learning}
C. Dong, C.~C. Loy, K. He, and X. Tang,
\newblock ``Learning a deep convolutional network for image super-resolution,''
\newblock in {\em European Conf. on Computer Vision}. 2014, pp. 184--199,
  Springer.

\bibitem{tiantong16deep}
T. Guo, H.~S. Mousavi, and V. Monga,
\newblock ``Deep learning based image super-resolution with coupled
  backpropagation,''
\newblock in {\em IEEE Global Conf. on Signal and Information Processing},
  2016, pp. 237--241.

\bibitem{bruna2015super}
J. Bruna, P. Sprechmann, and Y. LeCun,
\newblock ``Super-resolution with deep convolutional sufficient statistics,''
\newblock in {\em Int. Conf. on Learning Representations}, 2015.

\bibitem{dong2016image}
C. Dong, C.~C. Loy, K. He, and X. Tang,
\newblock ``Image super-resolution using deep convolutional networks,''
\newblock {\em IEEE Trans. on Pattern Analysis and Machine Intelligence}, vol.
  38, no. 2, pp. 295--307, 2016.

\bibitem{dong2016accelerating}
C. Dong, C.~C. Loy, and X. Tang,
\newblock ``Accelerating the super-resolution convolutional neural network,''
\newblock in {\em European Conf. on Computer Vision}. Springer, 2016, pp.
  391--407.

\bibitem{wang2015self}
Z. Wang, Y. Yang, Z. Wang, S. Chang, W. Han, J. Yang, and T.~S. Huang,
\newblock ``Self-tuned deep super resolution,''
\newblock {\em arXiv preprint arXiv:1504.05632}, 2015.

\bibitem{lai2017deep}
W.-S. Lai, J.-B. Huang, N. Ahuja, and M.-H. Yang,
\newblock ``Deep laplacian pyramid networks for fast and accurate
  super-resolution,''
\newblock in {\em IEEE Conf. on Computer Vision and Pattern Recognition}, 2017,
  pp. 624--632.

\bibitem{kim2016deeply}
J. Kim, J. Kwon~Lee, and K. Mu~Lee,
\newblock ``Deeply-recursive convolutional network for image
  super-resolution,''
\newblock in {\em IEEE Conf. on Computer Vision and Pattern Recognition}, 2016,
  pp. 1637--1645.

\bibitem{Kim_2016_VDSR}
J. Kim, J.~K. Lee, and K.~M. Lee,
\newblock ``Accurate image super-resolution using very deep convolutional
  networks,''
\newblock in {\em IEEE Conf. on Computer Vision and Pattern Recognition}, 2016,
  pp. 1646--1654.

\bibitem{guo2017deep}
T. Guo, H.~S. Mousavi, T.~H. Vu, and V. Monga,
\newblock ``Deep wavelet prediction for image super-resolution,''
\newblock in {\em IEEE Conf. on Computer Vision and Pattern Recognition
  Workshops}, 2017, pp. 1100--1109.

\bibitem{Junxuan2017}
S.~Y. Junxuan~Li and A. Robles-Kelly,
\newblock ``A frequency domain neural network for fast image
  super-resolution,''
\newblock {\em arXiv preprint arXiv: arXiv:1712.03037}, 2017.

\bibitem{zhao2003wavelet}
S. Zhao, H. Han, and S. Peng,
\newblock ``Wavelet-domain hmt-based image super-resolution,''
\newblock in {\em IEEE Int. Conf. on Image Processing}, 2003, pp. 933–--936.

\bibitem{robinson2010efficient}
M.~D. Robinson, C.~A. Toth, J.~Y. Lo, and S. Farsiu,
\newblock ``Efficient fourier-wavelet super-resolution,''
\newblock {\em IEEE Trans. on Image Processing}, vol. 19, no. 10, pp.
  2669--2681, 2010.

\bibitem{wahed2007image}
M.~E.-S. Wahed,
\newblock ``Image enhancement using second generation wavelet super
  resolution,''
\newblock {\em Int. Journal of Physical Sciences}, vol. 2, no. 6, pp. 149--158,
  2007.

\bibitem{ji2009robust}
H. Ji and C. Ferm{\"u}ller,
\newblock ``Robust wavelet-based super-resolution reconstruction: theory and
  algorithm,''
\newblock {\em IEEE Trans. on Pattern Analysis and Machine Intelligence}, vol.
  31, no. 4, pp. 649--660, 2009.

\bibitem{tiantong2018ortho}
T. Guo, H.~S. Mousavi, and V. Monga,
\newblock ``Orthogonally regularized deep networks for image
  super-resolution,''
\newblock in {\em IEEE Int. Conf. on Acoustics, Speech, and Signal Processing},
  2018, pp. 1463--1467.

\bibitem{bevilacqua2012low}
M. Bevilacqua, A. Roumy, C. Guillemot, and M.~L. Alberi-Morel,
\newblock ``Low-complexity single-image super-resolution based on nonnegative
  neighbor embedding,''
\newblock in {\em British Machine Vision Conf.}, 2012, pp. 1--10.

\bibitem{zeyde2010single}
R. Zeyde, M. Elad, and M. Protter,
\newblock ``On single image scale-up using sparse-representations,''
\newblock in {\em Int. Conf. on Curves and Surfaces}. Springer, 2010, pp.
  711--730.

\bibitem{MartinFTM01}
D. Martin, C. Fowlkes, D. Tal, and J. Malik,
\newblock ``A database of human segmented natural images and its application to
  evaluating segmentation algorithms and measuring ecological statistics,''
\newblock in {\em Int. Conf. on Computer Vision}, 2001, vol.~2, pp. 416--423.

\bibitem{huang2015single}
J.-B. Huang, A. Singh, and N. Ahuja,
\newblock ``Single image super-resolution from transformed self-exemplars,''
\newblock in {\em IEEE Conf. on Computer Vision and Pattern Recognition}, 2015,
  pp. 5197--5206.

\bibitem{yang2008image}
J. Yang, J. Wright, T. Huang, and Y. Ma,
\newblock ``Image super-resolution as sparse representation of raw image
  patches,''
\newblock in {\em IEEE Conf. on Computer Vision and Pattern Recognition}, 2008,
  pp. 1--8.

\bibitem{mousavi2017sparsity}
H.~S. Mousavi and V. Monga,
\newblock ``Sparsity-based color image super resolution via exploiting cross
  channel constraints,''
\newblock {\em IEEE Trans. on Image Processing}, vol. 26, no. 11, pp.
  5094--5106, 2017.

\bibitem{yang2010exploiting}
C.-Y. Yang, J.-B. Huang, and M.-H. Yang,
\newblock ``Exploiting self-similarities for single frame super-resolution,''
\newblock in {\em Asian Conf. on Computer Vision}. Springer, 2010, pp.
  497--510.

\bibitem{FreFat10}
G. Freedman and R. Fattal,
\newblock ``Image and video upscaling from local self-examples,''
\newblock {\em ACM Trans. Graph.}, vol. 28, no. 3, pp. 1--10, 2010.

\bibitem{nguyen2000efficient}
N. Nguyen and P. Milanfar,
\newblock ``An efficient wavelet-based algorithm for image superresolution,''
\newblock in {\em IEEE Int. Conf. on Image Processing}, 2000, vol.~2, pp.
  351--354.

\bibitem{jiji2004single}
C. Jiji, M.~V. Joshi, and S. Chaudhuri,
\newblock ``Single-frame image super-resolution using learned wavelet
  coefficients,''
\newblock {\em Int. journal of Imaging systems and Technology}, vol. 14, no. 3,
  pp. 105--112, 2004.

\bibitem{anbarjafari2010image}
G. Anbarjafari and H. Demirel,
\newblock ``Image super resolution based on interpolation of wavelet domain
  high frequency subbands and the spatial domain input image,''
\newblock {\em ETRI journal}, vol. 32, no. 3, pp. 390--394, 2010.

\bibitem{goodfellow2014generative}
I. Goodfellow, J. Pouget-Abadie, M. Mirza, B. Xu, D. Warde-Farley, S. Ozair, A.
  Courville, and Y. Bengio,
\newblock ``Generative adversarial nets,''
\newblock in {\em Advances in Neural Information Processing Systems}, 2014, pp.
  2672--2680.

\bibitem{johnson2016perceptual}
J. Johnson, A. Alahi, and L. Fei-Fei,
\newblock ``Perceptual losses for real-time style transfer and
  super-resolution,''
\newblock in {\em European Conf. on Computer Vision}. Springer, 2016, pp.
  694--711.

\bibitem{ledig2017photo}
C. Ledig, L. Theis, F. Husz{\'a}r, J. Caballero, et~al.,
\newblock ``Photo-realistic single image super-resolution using a generative
  adversarial network,''
\newblock in {\em IEEE Conf. on Computer Vision and Pattern Recognition}, 2017,
  vol.~2, pp. 105--114.

\bibitem{mousavi2018deep}
H.~S. Mousavi, T. Guo, and V. Monga,
\newblock ``Deep image super resolution via natural image priors,''
\newblock in {\em IEEE Int. Conf. on Acoustics, Speech, and Signal Processing},
  2018, pp. 1483--1487.

\bibitem{chen2018fsrnet}
Y. Chen, Y. Tai, X. Liu, C. Shen, and J. Yang,
\newblock ``Fsrnet: End-to-end learning face super-resolution with facial
  priors,''
\newblock in {\em IEEE Conf. on Computer Vision and Pattern Recognition}, 2018,
  pp. 2492--2501.

\bibitem{wang2018recovering}
X. Wang, K. Yu, C. Dong, and C.~C. Loy,
\newblock ``Recovering realistic texture in image super-resolution by deep
  spatial feature transform,''
\newblock in {\em IEEE Conf. on Computer Vision and Pattern Recognition}, 2018,
  pp. 606--615.

\bibitem{vishal2017image}
V. Monga,
\newblock {\em Handbook of Convex Optimization Methods in Imaging Science},
\newblock Springer, 2018.

\bibitem{he2012learning}
C. He, L. Liu, L. Xu, M. Liu, and M. Liao,
\newblock ``Learning based compressed sensing for sar image super-resolution,''
\newblock {\em Journal of Selected Topics in Applied Earth Observations and
  Remote Sensing}, vol. 5, no. 4, pp. 1272--1281, 2012.

\bibitem{wu2016super}
Z. Wu and H. Wang,
\newblock ``Super-resolution reconstruction of sar image based on non-local
  means denoising combined with bp neural network,''
\newblock {\em arXiv preprint arXiv:1612.04755}, 2016.

\bibitem{bahrami2016reconstruction}
K. Bahrami, F. Shi, X. Zong, H.~W. Shin, H. An, and D. Shen,
\newblock ``Reconstruction of 7t-like images from 3t mri,''
\newblock {\em IEEE Trans. on Medical Imaging}, vol. 35, no. 9, pp. 2085--2097,
  2016.

\bibitem{gonzalez1977digital}
R. Gonzalez and P. Wintz,
\newblock {\em Digital image processing},
\newblock Addison-Wesley Publishing Co., Inc., Reading, MA, 1977.

\bibitem{wallace1992jpeg}
G.~K. Wallace,
\newblock ``The jpeg still picture compression standard,''
\newblock {\em IEEE Trans. on Consumer Electronics}, vol. 38, no. 1, pp.
  xviii--xxxiv, 1992.

\bibitem{he2016identity}
K. He, X. Zhang, S. Ren, and J. Sun,
\newblock ``Identity mappings in deep residual networks,''
\newblock in {\em European Conf. on Computer Vision}. Springer, 2016, pp.
  630--645.

\bibitem{nair2010rectified}
V. Nair and G.~E. Hinton,
\newblock ``Rectified linear units improve restricted boltzmann machines,''
\newblock in {\em Int. Conf. on Machine Learning}, 2010, pp. 807--814.

\bibitem{noh2015learning}
H. Noh, S. Hong, and B. Han,
\newblock ``Learning deconvolution network for semantic segmentation,''
\newblock in {\em IEEE Int. Conf. on Computer Vision}, 2015, pp. 1520--1528.

\bibitem{dumoulin2016guide}
V. Dumoulin and F. Visin,
\newblock ``A guide to convolution arithmetic for deep learning,''
\newblock {\em arXiv preprint arXiv:1603.07285}, 2016.

\bibitem{techReport}
T. Guo, H.~S. Mousavi, and V. Monga,
\newblock ``A technical report on: orthogonally regularized deep networks for
  image super-resolution,'' 2017,
\newblock Code available on http://signal.ee.psu.edu/ORDSR.html.

\bibitem{probabilitybook}
K. Leonid and S.~Y. G.,
\newblock {\em Theory of Probability and Random Processes}, vol.~1,
\newblock Springer, 2007.

\bibitem{goodfellow2016deep}
I. Goodfellow, Y. Bengio, A. Courville, and Y. Bengio,
\newblock {\em Deep learning}, vol.~1,
\newblock MIT Press, 2016.

\bibitem{lecun1998gradient}
Y. LeCun, L. Bottou, Y. Bengio, and P. Haffner,
\newblock ``Gradient-based learning applied to document recognition,''
\newblock {\em Proceedings of the IEEE}, vol. 86, no. 11, pp. 2278--2324, 1998.

\bibitem{glorot2010understanding}
X. Glorot and Y. Bengio,
\newblock ``Understanding the difficulty of training deep feedforward neural
  networks,''
\newblock in {\em Int. Conf. on Artificial Intelligence and Statistics}, 2010,
  pp. 249--256.

\bibitem{Schulter_2015_CVPR}
S. Schulter, C. Leistner, and H. Bischof,
\newblock ``Fast and accurate image upscaling with super-resolution forests,''
\newblock in {\em IEEE Conf. on Computer Vision and Pattern Recognition}, 2015,
  pp. 3791--3799.

\bibitem{wang2004image}
Z. Wang, A.~C. Bovik, H.~R. Sheikh, and E.~P. Simoncelli,
\newblock ``Image quality assessment: from error visibility to structural
  similarity,''
\newblock {\em IEEE Trans. on Image Processing}, vol. 13, no. 4, pp. 600--612,
  2004.

\bibitem{sheikh2005information}
H.~R. Sheikh, A.~C. Bovik, and G. De~Veciana,
\newblock ``An information fidelity criterion for image quality assessment
  using natural scene statistics,''
\newblock {\em IEEE Trans. on Image Processing}, vol. 14, no. 12, pp.
  2117--2128, 2005.

\bibitem{deng2009imagenet}
J. Deng, W. Dong, R. Socher, L.-J. Li, K. Li, and L. Fei-Fei,
\newblock ``Imagenet: A large-scale hierarchical image database,''
\newblock in {\em IEEE Conf. on Computer Vision and Pattern Recognition}. IEEE,
  2009, pp. 248--255.

\bibitem{lin2014microsoft}
T.-Y. Lin, M. Maire, S. Belongie, and orthers,
\newblock ``Microsoft coco: Common objects in context,''
\newblock in {\em European Conf. on Computer Vision}. Springer, 2014, pp.
  740--755.

\bibitem{dong2011image}
W. Dong, L. Zhang, G. Shi, and X. Wu,
\newblock ``Image deblurring and supe r-resolution by adaptive sparse domain
  selection and adaptive regularization,''
\newblock {\em IEEE Trans. on Image Processing}, vol. 20, no. 7, pp.
  1838--1857, 2011.

\bibitem{tensorflow2015-whitepaper}
M. Abadi, A. Agarwal, and P.~B. et. al.,
\newblock ``{TensorFlow}: Large-scale machine learning on heterogeneous
  systems,''
\newblock {\em arXiv preprint arXiv:1603.04467}, 2015.

\bibitem{timofte2014a+}
R. Timofte, V. De~Smet, and L. Van~Gool,
\newblock ``A+: Adjusted anchored neighborhood regression for fast
  super-resolution,''
\newblock in {\em Asian Conf. on Computer Vision}. 2014, pp. 111--126,
  Springer.

\bibitem{wang2015deep}
Z. Wang, D. Liu, J. Yang, W. Han, and T. Huang,
\newblock ``Deep networks for image super-resolution with sparse prior,''
\newblock in {\em IEEE Int. Conf. on Computer Vision}, 2015, pp. 370--378.

\bibitem{zhang2018residual}
Y. Zhang, Y. Tian, Y. Kong, B. Zhong, and Y. Fu,
\newblock ``Residual dense network for image super-resolution,''
\newblock in {\em IEEE Conf. on Computer Vision and Pattern Recognition}, 2018,
  pp. 2472--2481.

\bibitem{lim2017enhanced}
B. Lim, S. Son, H. Kim, S. Nah, and K.~M. Lee,
\newblock ``Enhanced deep residual networks for single image
  super-resolution,''
\newblock in {\em IEEE Conf. on Computer Vision and Pattern Recognition
  Workshop}, 2017, pp. 1132--1140.

\bibitem{timofte2013anchored}
R. Timofte, V. De~Smet, and L. Van~Gool,
\newblock ``Anchored neighborhood regression for fast example-based
  super-resolution,''
\newblock in {\em IEEE Int. Conf. on Computer Vision}, 2013, pp. 1920--1927.

\bibitem{tai2017image}
Y. Tai, J. Yang, and X. Liu,
\newblock ``Image super-resolution via deep recursive residual network,''
\newblock in {\em IEEE Conf. on Computer Vision and Pattern Recognition}, 2017,
  pp. 2790--2798.

\bibitem{timofte2016seven}
R. Timofte, R. Rothe, and L. Van~Gool,
\newblock ``Seven ways to improve example-based single image super
  resolution,''
\newblock in {\em IEEE Conf. on Computer Vision and Pattern Recognition}, 2016,
  pp. 1865--1873.

\end{thebibliography}
\end{document}